\documentclass{article}

\usepackage[preprint]{neurips_2024}

\usepackage[utf8]{inputenc} 
\usepackage[T1]{fontenc}    
\usepackage{hyperref}       
\usepackage{url}            
\usepackage{booktabs}       
\usepackage{amsfonts}       
\usepackage{nicefrac}       
\usepackage{microtype}      
\usepackage{xcolor}         

\usepackage{microtype}
\usepackage{graphicx}
\usepackage{subfigure}
\usepackage{booktabs} 
\usepackage{amsmath}

\newcommand{\vve}[1]{\mathbf{#1}}
\usepackage{amssymb}
\usepackage{mathtools}
\usepackage{amsthm}
\usepackage{wrapfig}

\title{Gradient-based inference of abstract task representations for generalization in neural networks}

\author{%
  Ali Hummos\\
  Brain and Cognitive Sciences Department\\
  Massachusetts Institute of Technology\\
  Cambridge, MA\\
  \texttt{ahummos@MIT.edu} \\
  \And
  Felipe del Río \\
  Department of Computer Science\\
  Pontificia Universidad Catolica de Chile\\
  Santiago, Chile\\
  \And
  Brabeeba Mien Wang\\
  Computer Science and Artificial Intelligence Laboratory\\
  Massachusetts Institute of Technology\\
  Cambridge, MA\\
  \And
  Julio Hurtado\\
  Centre for Applications of Mathematical \& Computing Sciences\\
  University of Warwick\\
  \And
  Cristian B. Calderon\\
  Centro Nacional de Inteligencia Artificial\\
  Santiago, Chile\\
  \And
  Guangyu Robert Yang \\
  Brain and Cognitive Sciences Department\\
  Massachusetts Institute of Technology\\
  Cambridge, MA\\
}

\begin{document}

\maketitle

\begin{abstract}
Humans and many animals show remarkably adaptive behavior and can respond differently to the same input depending on their internal goals. The brain not only represents the intermediate abstractions needed to perform a computation but also actively maintains a representation of the computation itself (task abstraction). Such separation of the computation and its abstraction is associated with faster learning, flexible decision-making, and broad generalization capacity. We investigate if such benefits might extend to neural networks trained with task abstractions. For such benefits to emerge, one needs a task inference mechanism that possesses two crucial abilities: First, the ability to infer abstract task representations when no longer explicitly provided (task inference), and second, manipulate task representations to adapt to novel problems (task recomposition). To tackle this, we cast task inference as an optimization problem from a variational inference perspective and ground our approach in an expectation-maximization framework. We show that gradients backpropagated through a neural network to a task representation layer are an efficient heuristic to infer current task demands, a process we refer to as gradient-based inference (GBI). Further iterative optimization of the task representation layer allows for recomposing abstractions to adapt to novel situations. Using a toy example, a novel image classifier, and a language model, we demonstrate that GBI provides higher learning efficiency and generalization to novel tasks and limits forgetting. Moreover, we show that GBI has unique advantages such as preserving information for uncertainty estimation and detecting out-of-distribution samples.
\end{abstract}

\section{Introduction}


Cognitive science and neuroscience hold a prominent place for (top-down) abstract task representations in many accounts of advanced cognitive functions in animals, including humans \citep{niv_learning_2019}. The brain not only represents the intermediate abstractions needed to perform a task but also actively maintains a representation of the task itself (i.e., task abstraction, \citep{mante_context-dependent_2013, rikhye_thalamic_2018, zhou_complementary_2019, vaidya_neural_2021, hummos_thalamic_2022}). Such separation of the computation and its abstraction is theorized to support learning efficiency, as well as adaptive behavior. First, regarding learning efficiency, task abstractions facilitate learning by organizing experiences collected \citep{yu_adaptive_2021}. Human participants learn faster once they discover the underlying task structure \citep{badre_frontal_2010, collins_cost_2017,vaidya_neural_2021, castanon_mixture_2021}. In fact, humans have a bias to discover and use such latent task structures even when they do not exist or might hurt their performance \citep{gaissmaier_smart_2008, collins_cost_2017}. Second, regarding adaptive behavior, previous work has shown that the brain updates these task abstractions to meet current task demands of the environment and (re)composes them to solve new problems \citep{miller_integrative_2001, collins_reasoning_2012}. Indeed, depending on the context, the brain is able to respond to the same sensory input in novel ways by composing an appropriate task abstraction that guides processing of the input \citep{rikhye_thalamic_2018, tafazoli_building_2024}, supporting flexible decision making and generalization to novel situations \citep{collins_reasoning_2012, vaidya_neural_2021}.

Traditional artificial neural networks (ANNs) architectures have a static computational graph that entangles input processing with task inference. For task-dependent computations in ANNs, one popular solution is to build in task representations in the networks. One solution is to build task representations into the structure of the network by using modular networks; where each module could represent a given task \citep{andreas_neural_2016, kirsch_modular_2018, goyal_recurrent_2019}. Alternatively, one could regularize the weight updates such as to maximize an orthogonal computational space for each task \citep{kirkpatrick_overcoming_2017, masse_alleviating_2018}. Our framework is most related to models that add task encoding input layers, which provide information about the current task demands to the network \citep{yang_task_2019, hurtado_optimizing_2021, wang_contextual_2022}.

However, these previous models lack two important task inference features which we unpack in what follows and motivate our work. First, these models do not propose a mechanism to efficiently identify these tasks if they appear again in data, and thereby flexibly (re)adapt to the demands of previously encountered task\footnote{This work sidesteps the challenge of unsupervised discovery of task abstractions, a problem tackled by several algorithms \citep{butz_learning_2019, xie_deep_2021, mazzaglia_choreographer_2022, hummos_thalamus_2023}}. Second, and perhaps most importantly, these models lack a principled way of recomposing previously learned task abstractions, thereby minimizing the necessity for new learning (i.e. parameters update) to adapt to new situations \citep{lake_building_2017}. 


To address the challenge of an efficient inference mechanism that can capture the two properties just described, we propose Gradient-Based Inference (GBI) of task abstractions, a solution grounded in variational inference, offering a robust and flexible framework. While traditional variational models have employed a static encoder to map data to an approximate posterior over latent representations (e.g. \citep{kingma_auto-encoding_2013}), they also have the potential to iteratively refine the posterior through rounds of optimization along gradients through the decoder \citep{rezende_stochastic_2014}. Previous models have used iterative optimization in latent space to discover latent tasks and associate them with an internally generated latent task embedding \citep{butz_learning_2019, hummos_thalamus_2023}. While these models were able to generalize and learn continually, the iterative refinement can be computationally demanding. GBI tackles the computational efficiency by investigating a role for the decoder gradients in lieu of likelihoods and reformulating them in probabilistic terms. We show that one-step gradient updates can provide an informative distribution that has sufficient accuracy and can support detecting OOD samples and estimating uncertainty.


Our findings confirm that empirical findings from cognitive science do indeed extend to neural networks. In experiment 1 (toy dataset, section \ref{section:bayesian_task}, we show that ANN models provided with abstract task representations do in fact show faster learning (improved data-efficiency), improved generalization to novel data points, and reduced forgetting. Importantly, we also confirm that a neural network trained with these abstractions has all the information to infer them at test time through back-propagated gradients, acting as a generative model that can estimate the likelihood of each abstraction and how well it explains the input data. In experiment 2 (image classification, section \ref{section:classification}), we first demonstrate that GBI can be used to perform image classification in a image generation setting. Second, we show that on-step gradient updates show interesting properties such as conveying uncertainty information and allowing for the detection of out-of-distribution (OOD) samples. Third, we consider a larger dataset with more categories using CIFAR-100. Finally, in experiment 3 (language experiment, section \ref{section:language}), we show that GBI is a domain-general method, and demonstrate several GBI properties (i.e., task inference, data efficiency, generalization) in language modeling.

\section{Methods}
\paragraph{Overview.} We assume a dataset is equipped with a task abstraction. Conceptually, a task abstraction groups data samples into conceptual categories defined across dataset samples rather than individual data points. Mathematically, we model this by a graphical model where the data point $\mathbf{X}$ is generated from a task abstraction $\mathbf{Z}$. Now we use a neural network to model this graphical model with data $\mathbf{X}$, task abstraction $\mathbf{Z}$ and unknown parameter $\theta$, our neural network model has the form of a joint likelihood function
\begin{equation}
    \mathcal{L}(\theta; \mathbf{X}, \mathbf{Z}) = p(\mathbf{X},\mathbf{Z} | \theta).
\end{equation}

At the training phase, given a data point $\mathbf{X}$ with the abstraction of interest $\hat{\mathbf{Z}}$ directly observed, we train the neural network by doing gradient descent on $-\log\mathcal{L}(\theta; \mathbf{X}, \hat{\mathbf{Z}})$ with respect to $\theta$. 

At the test phase, we no longer assume access to task abstraction $\mathbf{Z}$ and require the model to identify them in individual data points efficiently. Moreover, we require for our model a mechanism to manipulate the abstract representations to adapt to new unseen situations. Specifically, we update our $\mathbf{Z}$ through gradient descent on $-\log\mathcal{L}(\theta; \mathbf{X}, \mathbf{Z})$.

The prediction of the model with task abstraction $\mathbf{Z}$ is
\begin{equation}
    \hat{\mathbf{X}} = \arg\max_{\mathbf{X}} \mathcal{L}(\theta; \mathbf{X}, \mathbf{Z}).
\end{equation}

\paragraph{Connections to Expectation-Maximization algorithm.} Notice that this method has deep connections to the classical Expectation-Maxmimization (EM) algorithm. In the training phase, since $\hat{\mathbf{Z}}$ is drawn from the distribution $p_{\mathbf{Z} | \mathbf{X}}$, the training attempts to find $\hat{\theta}$ such that
\begin{equation}
\label{eq: em compact}
    \hat{\theta} = \arg\max_\theta\mathbb{E}_{p_{\mathbf{Z} | \mathbf{X}}}[\log\mathcal{L}(\theta; \mathbf{X}, \mathbf{Z})].
\end{equation}
\autoref{eq: em compact} is a compact way to write the classical EM algorithm where the expectation steps are replaced by direct observation. Let the probability distribution of $\mathbf{Z}$ be $q$. We can write
\begin{align}
    \log p(\mathbf{X} | \theta) &= \mathbb{E}_q[\log p(\mathbf{X} | \theta)]\\
    &= \mathbb{E}_q[\log p(\mathbf{X}, \mathbf{Z} | \theta) - \log p(\mathbf{Z} | \mathbf{X}, \theta)].\\
    \intertext{By adding and subtracting an entropy of $q$, we have}
    &= \mathbb{E}_q[\log p(\mathbf{X}, \mathbf{Z} | \theta)] + H(q) + KL(q || p_{\mathbf{Z} | \mathbf{X}})
\end{align}
where KL divergence $KL(q || p)$ is defined as $-\mathbb{E}_q[\log\frac{p}{q}]$ and the entropy $H(q)$ is defined as $-\mathbb{E}_q[\log q]$. Now the expectation step corresponds to find $q$ such that $\mathbb{E}_q[\log p(\mathbf{X}, \mathbf{Z} | \theta)] + H(q)$ is maximized. Notice that this is equivalent to minimizing the KL divergence $KL(q || p_{\mathbf{Z} | \mathbf{X}})$ and we know it is minimized at $0$ when $q = p_{\mathbf{Z} | \mathbf{X}}$. In particular, we have $\mathbb{E}_q[\log p(\mathbf{X}, \mathbf{Z} | \theta)] + H(q) =   \log p(\mathbf{X} | \theta)$.

At the maximization step, we find $\theta$ such that $ \mathbb{E}_q[\log p(\mathbf{X}, \mathbf{Z} | \theta)] + H(q)$ is maximized, but since $H(q)$ does not depend on $\theta$, it is equivalent to maximize $\mathbb{E}_q[\log p(\mathbf{X}, \mathbf{Z} | \theta)]$ and therefore it is exactly \autoref{eq: em compact}.

At our testing phase, since we are not able to access $p_{\mathbf{Z} | \mathbf{X}}$ directly, we optimize $q$ through gradient descent. Our objective is 
exactly the expectation step with the regularization term $H(q)$ implemented as L2 regularization on $\mathbf{Z}$. 

There are different methods to optimize $\mathbf{Z}$ and we discuss two methods we use in this paper, iterative optimization and one-step gradient update below to obtain gradient-based inference (GBI) estimates over latent variable values. 

\paragraph{Iterative optimization of $\mathbf{Z}$.}  One method for doing gradient descent on $\mathbf{Z}$ is to iteratively optimize $\vve{z}$ using the gradient ${\partial\mathcal{L}} / {\partial\vve{z}}$. We implement iterative optimization using Adam optimizer with a learning rate of 0.01 and L2 regularization scaled by 0.01. This method usually takes many iterations \citep{marino_iterative_2018}. Another method is to learn an update rule $f_{\phi}(\cdot)$ that converts the gradient directly into an inferred $\vve{z} \leftarrow f_{\phi}({\partial\mathcal{L}} / {\partial\vve{z}})$ \citep{marino_iterative_2018}. This method allows for rapid inference during run time but requires training of the update rule (like other meta-learning methods). 

\paragraph{One-step gradient update.} We explore an alternative that requires only one pass through the model by updating the gradient only once. Without iteratively optimizing $\mathbf{Z}$ through gradient update, one-step gradient update usually heavily depends on the initialization: because if the initialization is far away from the optimal $\mathbf{Z}$, the local gradient information might not be informative of the global optimum. However, notice that in the expectation step, our goal is to find the probability distribution of $\mathbf{Z}$, $q$, such that $\mathbb{E}_q[\log p(\mathbf{X}, \mathbf{Z} | \theta)] + H(q)$ is maximized. We can consider an alternative objective function $f_{\alpha}(q) = \mathbb{E}_q[\log p(\mathbf{X}, \mathbf{Z} | \theta)] + \alpha H(q)$
 for some $\alpha > 0$ and let $q^*_{\alpha}$ be the point that maximizes this alternative objective function. Since entropy is concave, by increasing $\alpha$ we make the function more concave and therefore the gradient is more informative of the actual optimum. Furthermore, $q^*_{\alpha}$ will also become closer to the maximal entropy point $\bar{q} = \arg\max_{q} H(q)$. Therefore we can choose our initialization point as the maximal entropy point and output $\hat{q}$ using our one step gradient update rule:
\begin{equation}
\hat{q} = \text{softmax}(\bar{q} + \nabla f_{\alpha}(\bar{q})).
\end{equation}
Here, we use a softmax function to project $q$ back to the space of distribution. We implement this update using the SGD optimizer with learning rate 1 and no momentum terms. We reset our initialization point to maximal entropy to obtain $\hat{q}$ for the next data point.

Code available at \url{https://anonymous.4open.science/r/neuralbayes-F06F/}.

\begin{figure}[h]
  \centering 
  \includegraphics[width=0.80\textwidth]{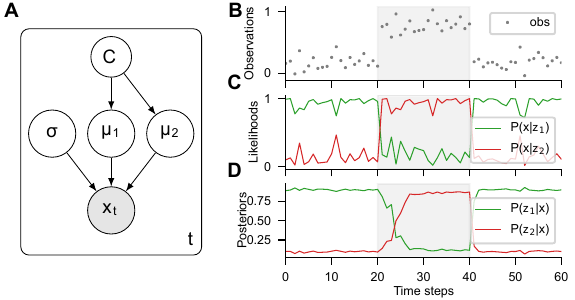}
  \caption{Sequence prediction task in one dimension. A) Bayesian graphical model generating the data. Task node (z) flips periodically and sets the mean of the Gaussian generating the data between $\mu_1$ = 0.2 or $\mu_2$ = 0.8, with fixed standard deviation $\sigma$ at 0.1. B) Sample generated data observations. C) The likelihood of each context node value given data is calculated using the Gaussian probability density equation, more details in Appendix B. D) The posterior over z values.}
  \label{fig:toy_task}
\end{figure}

\section{Experiments}
The experiments are organized as follows. In Section \ref{section:bayesian_task}, we first investigate the properties of GBI on a simple synthetic dataset generated from a ground truth Bayesian model. We show that GBI displays better data-efficiency, generalizes better and forgets less, in addition to being able to pass neural gradients back to the Bayesian model to support Bayesian computations. Section \ref{section:classification} assesses the potential of GBI as a novel image classifier, the GBI ability to generate task samples, and to detect OOD samples. Finally, in Section \ref{section:language}, we demonstrate the abilities of GBI to recompose previously learned tasks, and thereby flexibly adapt to novel tasks using a language model.


\subsection{Experiment 1: The impact of task abstractions and the dynamics of GBI in a toy dataset}
\label{section:bayesian_task}

We begin with a simple dataset of one-dimensional observations unfolding in time, $\vve{x}_{0:t}$. The observations are generated from two alternating Gaussian distributions with distinct means ($\mathcal{N}(0.2, 0.1)$ or $\mathcal{N}(0.8, 0.1)$, Fig \ref{fig:toy_task}A, B). The ground truth generative causal model (Fig \ref{fig:toy_task}A) has the task node ($\vve{z}$) as\ a binary variable with values either $\mathbf{z^1}$ = [0, 1] or $\mathbf{z^2}$ = [1, 0]. Context node switched between those two values with a probability $P_v$ = 0.005, but we enforced a minimum block length of 20 and a maximum of 50 (further details in Appendix B). Knowing the ground truth generative model, we can analytically calculate the likelihood of each $z^i$ as $p(x_t | z^i)$ and the posteriors $p(z^i | x_{0:t})$(Fig \ref{fig:toy_task}C-E). 

To estimate these Bayesian quantities from a neural network, we train a 100-unit LSTM \citep{hochreiter_long_1997} to predict the next observation $x_{t+1}$ given: (i) the five previous observations (\{$x_{t},...,x_{t-h}$\}, h = 5) presented sequentially, and (ii) a task abstraction input which we set to the task z values, one-hot encoded with either [0, 1] or [1, 0] after passing through a softmax function. We name this model GBI-LSTM and train it on the task (training details in appendix B), and in the following sections, we compare its learning dynamics to an LSTM with no such task input. 

\begin{figure*}[h]
  \centering
  \includegraphics[width=\textwidth]{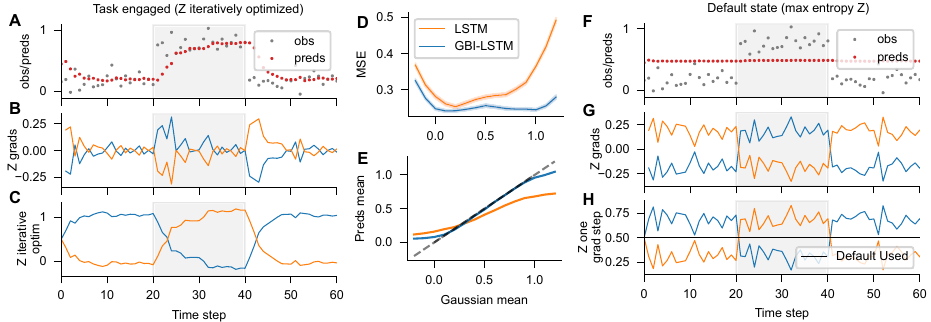}
  \caption{Neural network trained with task abstractions can dynamically tune itself to incoming data and generalizes to novel data points. A) Responses of the GBI-LSTM to sample data and B) the gradients of the prediction loss w.r.t. to the two Z units. C) Iteratively optimized Z values accumulating evidence dynamically. D) As a generalization test, while the network was trained on data from Gaussians with means 0.2 and 0.8, we test its responses to Gaussians between -0.2 to 1.2 (in 0.1 increments). We record the mean MSE values across 20 runs with different random seeds and compare to the baseline LSTM trained without task abstraction input.  We only show runs where the last block of training was a 0.2 block, to highlight models behavior on that mean vs the other (0.8). E) The mean of the Gaussian data vs. the mean network predictions in each generalization block. These results are further quantified in Table \ref{table:toy_task_quantification_of_losses}. F-H) We show the other mode of using the same GBI-LSTM for gradient-based inference. We fix the task abstraction input Z to its state of maximum entropy (here [0.5, 0.5]) and take gradients from that point. While the network responds in the middle (F), the gradients w.r.t to Z (G) or similarly, the one-step gradient descent Z values behave much like the likelihood function, c.f. values computed using the Gaussian PDF in Fig \ref{fig:toy_task}C.
  }
  \label{fig:toy_task_GBI_results}
\end{figure*}

\paragraph*{GBI-LSTM is data efficient.} GBI-LSTM learns the task faster (i.e. better data efficiency) (Fig \ref{fig:toy_task_training_losses}A,B). Example runs in appendix B (Fig S\ref{fig:bayesian_lstm_with_context}), (LSTM, Fig S\ref{fig:bayesian_lstm_no_context}).
On one hand, this seems obvious as the GBI-LSTM gets additional contextual information, but on the other hand, the task is simple enough that it is not clear that a 100-unit LSTM needs the additional information. To explain, we reasoned that the available task abstraction deeply influences what solutions the network learns and allows for modules to emerge for each task. We used `task variance' to identify neurons active in each task, a measure used in neuroscience literature \citep{yang_task_2019}. We identify active neurons by selecting units with the highest variance across samples of inputs and outputs for a task. We see that neurons for each task overlap in the LSTM but separate into modules in the GBI-LSTM, throughout training (Fig \ref{fig:toy_task_training_losses}C). At \textit{inference time}, with weights frozen, the baseline LSTM handles task transitions efficiently (Fig S\ref{fig:bayesian_lstm_no_context}). The GBI-LSTM has the weights frozen but we now iteratively optimize the task abstraction layer z using vanilla SGD optimizer with a learning rate of 0.5 and we see that it also transitions between tasks by updating its task abstraction input z (Fig \ref{fig:toy_task_GBI_results}A-C, Fig S\ref{fig:bayesian_lstm_with_context}). 

\begin{figure}
    \centering
    \includegraphics[width=0.65\textwidth]{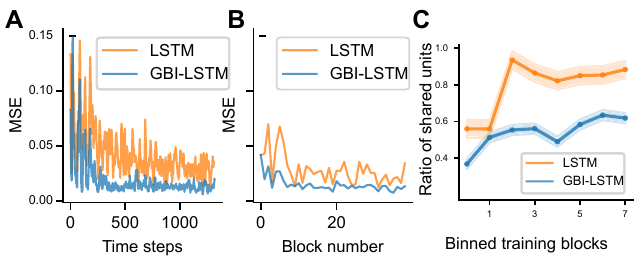}
    \caption{Comparing the training loss of an LSTM trained with task input (GBI-LSTM) without task input (LSTM). Note the variability in these loss estimates despite the simulation run over 10 random seeds is due to the stochastic block transitions still coinciding frequently. B) We show the mean loss per block but exclude the initial 20 predictions in each block to exclude poor performance from either model limited to transitioning between tasks. C) We quantify the ratio of shared neurons active in the 0.2 blocks and the 0.8 blocks using `task variance' to identify engaged neurons. We binned training blocks into groups of 5 to aggregate the data and increase accuracy.}
    \label{fig:toy_task_training_losses}
\end{figure}

\paragraph*{GBI-LSTM generalizes better and forgets less.} GBI-LSTM generalizes better to values outside of the training range (Fig \ref{fig:toy_task_GBI_results}D). By using iterative optimization, a gradient step in $\mathbf{Z}$ every time step, the GBI-LSTM can interpolate and better predict data points drawn from other distributions far from the two training distributions (Fig \ref{fig:toy_task_GBI_results}E). 
Moreover, we observe that the baseline LSTM already shows signs of catastrophic forgetting in this very simple setting. 
Testing MSE is worse around one of the two training means depending on which mean generated the last block of data during training. In figure \ref{fig:toy_task_GBI_results}D, we show the responses from runs where the last training block had the 0.2 Gaussian mean active. In contrast, as quantified in Table \ref{table:toy_task_quantification_of_losses} the GBI-LSTM shows no signs such forgetting (Fig \ref{fig:toy_task_GBI_results}E).

\begin{table}[h]
    \centering
        \caption{Quantifying the generalization and forgetting on the toy task. Both models were trained on data generating means 0.2 and 0.8. We grouped models prediction errors on data generated from means -0.2 through 1.2 into four informative categories and evaluated mean MSE values $\pm$ SEM. We identified the data generating mean during the last block of the training sequence, as the LSTM overfits to this mean and forgets data from the other mean. 20 runs with different seeds.}
    \begin{tabular}{lcc}
        \hline
        \textbf{Data range} & \textbf{LSTM MSE} & \textbf{GBI-LSTM MSE}\\
        \hline
        Mean of the last training block (0.2 or 0.8) & 0.25 $\pm$ 0.02 & \textbf{0.24} $\pm$ 0.02 \\
        The other mean (0.2 or 0.8)& 0.30 $\pm$ 0.03 & \textbf{0.24} $\pm$ 0.02 \\
        Inside training range (0.3-0.7)& 0.27 $\pm$ 0.02 & \textbf{0.25} $\pm$ 0.01 \\
        Outside training range ($<$0.2 \& $>$0.8) & 0.35 $\pm$ 0.06 & \textbf{0.26} $\pm$ 0.03 \\
        \hline
    \end{tabular}

     \label{table:toy_task_quantification_of_losses}
\end{table}

\paragraph*{Bayesian properties of GBI-LSTM.} We can use our GBI-LSTM for inference, i.e. approximate Bayesian quantities such as the posterior and the likelihood function of z. We distinguish between two ways we can use the GBI-LSTM, first by taking one-step gradient updates from maximal entropy points (`default state' mode) and second is the iterative optimization of the task abstraction z at a lower learning rate (`task-engaged' mode) which we used above to perform the task and generalize to novel data. In the `default state' mode, we set the task abstraction layer to its state of maximum entropy (here, the mean of the two values observed in training). For each time step, we take one step in z space to lower the prediction error, and then reset to max entropy for the next point. 
These one step z values (Fig \ref{fig:toy_task_GBI_results}H) behave much like the likelihoods of z computed through the ground truth causal model (Fig \ref{fig:toy_task}C). (See Fig S\ref{fig:overlaid_gradients_and_likelihoods} for a plot of the likelihoods and one gradient step z overlaid). Moreover, we see that these one step z can support Bayesian computations when used in lieu of the likelihood term in the Bayesian equation (previously calculated using the Gaussian probability distribution equation). 
i.e., we are able to pass one step z values to a Bayesian graphical model and Bayesian computations proceed normally (Fig S\ref{fig:bayesian_gradients_to_bayesian}). As such, the Bayesian model was able to offload learning parts of the causal graph to the neural network, but maintained the ability to pass belief updates and messages across the directed graph, despite the grafting of a neural network in the graph. 


\begin{wraptable}{R}{0.5\textwidth} 
        \centering
        \caption{GBI outperforms other gradient-based inference methods, and compares well to other canonical methods in ML (small drop in accuracy, but only one run through the neural network component). For iterative optimization, we further examine the relationship between optimization steps and accuracy in Fig S\ref{fig:vi-performance-per-steps}.}
    \label{table:accuracies_mnist}
        \resizebox{.50\textwidth}{!}{
            \begin{tabular}{l c r}
            \hline
            \textbf{Method} & \textbf{Accuracy (\%)}  &  \textbf{Runs} \\
            \hline
            \textbf{Canonical methods} & & \\
            Classifier & 91.44 $\pm$ 0.51 & 1 \\
            Pure generative & 81.91 $\pm$ 2.3 & 10 \\
            \hline
            \textbf{Ours} & & \\
            GBI & 85.46 $\pm$ 3.60 &  1 \\
            Iterative Optimization & 88.51 $\pm$ 3.42 & 50 \\
            Likelihood (Recon MSE) & 91.94 $\pm$ 3.38 &  10 \\
            \hline
            \textbf{Other gradient-based methods} & & \\
            EBI & 27.37 $\pm$ 2.22 & 1 \\
            NBI & 78.78 $\pm$ 1.21 &  10 \\
            \hline
            \end{tabular}
        }
\end{wraptable}
\subsection{Experiment 2: GBI properties for image classification}
\label{section:classification}
\begin{figure*}[]
  \centering
  \includegraphics[width=.7\textwidth]
  {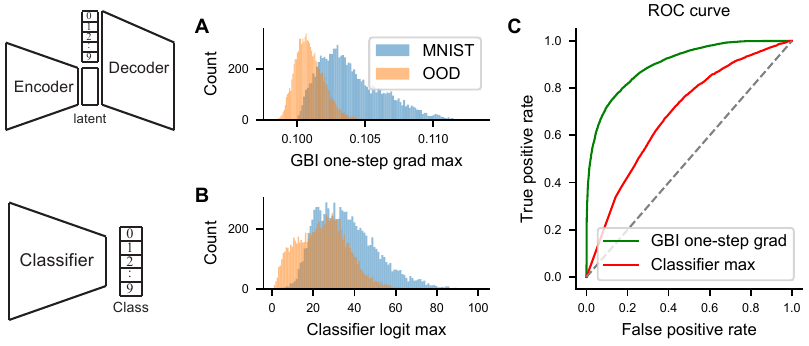}
  \caption{Robust out-of-distribution (OOD) samples detection in a GBI autoencoder compared to a classifier. Both models were trained on MNIST digits, and tested on OOD dataset fashionMNIST. We take the A) max value of the task abstraction layer in GBI and for the B) logits in the classifier and consider how are they distributed for in-distribution samples and OOD samples. C) ROC curves show that GBI max values are more discriminative.}
  \label{fig:ood_distributions}
\end{figure*}

We next explore the properties of gradient-based inference in more complex datasets. We trained a convolutional autoencoder to reconstruct MNIST digits. The decoder gets an additional input of a one-hot encoding of the digit label in addition to the latent embedding from the encoder (Fig \ref{fig:ood_distributions} Schematic). We train this structure for 4 epochs using an MSE reconstruction loss (full details in Appendix C). Conforming the results observed in the toy task, the network receiving additional digit label input trains faster compared with baseline autoencoder,  (Fig S\ref{fig:losses-gbi-vs-ae}).

\paragraph*{Gradients infer image class.} At test time, we set the task abstraction (i.e., digit classes) input z to maximum entropy point and take one gradient step of the reconstruction loss with respect to the class inputs z, and directly interpret the resulting vector to infer the digit label. We make two kinds of comparison, first to methods that used gradient-based inference, we refer to them as entropy-based \citep{wortsman_supermasks_2020} and norm-based inference\citep{roy_efficient_2024} and we see that GBI outperforms previous gradient-based methods (Table \ref{table:accuracies_mnist}, implementation details in appendix C). Second, we compare GBI with four canonical methods to infer class based on the same architecture, as follows: \textbf{(i)} a convolutional classifier of the same architecture as the encoder, as a measure of what one might gain from a dedicated inference model that maps from inputs to task abstraction (in this case, digit class). \textbf{(ii)} iteratively optimizing the latent following the traditional variational inference paradigm (VI), which is the same as our model, but increases the computational budget by taking many smaller steps of gradient descent in task abstraction space. \textbf{(iii)} evaluating the model's MSE loss at each digit one-hot encoding and selecting the one with lowest MSE loss, as a general representation of Bayesian methods that require evaluating a likelihood function over all hypotheses of interest. \textbf{(iv)} evaluating the model's reconstruction MSE again, but in a purely generative model trained from task abstraction input to image reconstruction with no encoder, as an ablation to the separation of task input and task abstraction in our model. The goal: show that GBI suffers only a small drop in accuracy while offering a small computational budget and no added parameters. We see that GBI outperforms other gradient-based methods EBI and NBI, and show only a small drop in accuracy compared to canonical methods in machine learning (Table \ref{table:accuracies_mnist}). We can also use our same model and increase the computational budget with iterative optimization or serial likelihood assessments when higher accuracy is required. More importantly, GBI offers a set of unique advantages as we discuss next.

\paragraph*{Advantages of GBI in image classification.} GBI retains many of the desirable properties of a generative model and thus can evaluate the probability of the data under the model. Therefore, we reasoned that GBI may have an advantage in uncertainty estimation (Fig S\ref{fig:confidence_GBI_vs_classifier}) and in detecting out-of-distribution (OOD) over traditional models. To test this, We compared a traditional convolutional classifier with GBI, both trained on MNIST digits and using fashionMNIST clothing items as the OOD dataset \citep{xiao_fashion-mnist_2017}. Using one gradient step task abstraction layer logits, GBI is superior at detecting OOD samples compared with the convolutional classifier (figure\ref{fig:ood_distributions}). In comparing our OOD detection results to current state-of-the-art methods we noted that the standard setup in studying OOD detection does not involve normalizing the image brightness mean and variance, making it trivial to discriminate the in and out-of-distribution datasets. We compared our OOD results to the Likelihood Regret (LR) method for OOD detection in a generative model \cite{xiao_likelihood_2020} (see supplementary material for implementation details). On unnormalized data samples, where detection methods can take advantage of differences in pixel intensities between the in and out-of-distribution samples, both GBI and LR perform high (Fig A\ref{fig:ood_not_normalized}, Table A\ref{table:OOD_comparison_unnormalized}). However, when data samples are normalized GBI shows a clear advantage over LR (table \ref{table:OOD_comparison}). Lastly, we provide an illustration of how the GBI framework enables adaptive behavior, where in response to the same input (image), changing the task abstraction (the digit label) can lead to a different input-to-output transformation, allowing for an additional layer of flexibility in task selection (Fig S\ref{fig:conditional_generation}).

\begin{wraptable}{R}{0.5\textwidth} 
\centering
\caption{One gradient step values in GBI show better OOD detection compared to classifier logit and state-of-the-art method. AUCROC values averaged over 10 seeds with standard deviations reported. We examine the case when pixel intensity statistics were normalized so methods are not trivially detecting changes in brightness between the in-distribution and OOD datasets. Unnormalized results in Tab S\ref{table:OOD_comparison_unnormalized}}
\label{table:OOD_comparison}
\resizebox{.47\textwidth}{!}{
\begin{tabular}{l c}
\hline
\textbf{Method} & \textbf{AUCROC} \\
\hline
GBI & \textbf{0.89} $\pm$ 0.03 \\
Classifier Softmax Maxes & 0.73 $\pm$ 0.08 \\
Likelihood Regret & 0.80 $\pm$ 0.06 \\
\hline
\end{tabular}
}

\end{wraptable}

\paragraph{CIFAR-100.} We apply GBI to the CIFAR-100 dataset, which features color images of vehicles, animals with diverse backgrounds. The GBI accuracy 18\% well above that obtained from a pure generative model (Table \ref{table:cifar100-methods-results}) (i.e. a model to decode image class to image directly). This suggests a positive effect from the separation of visual input and task abstraction streams. We see only a small drop between evaluating the likelihood of each image class for each image (requires one for each of the 100 image classes, accuracy 21\%) and GBI which requires only 1 run. Of note, we switched from an additive task abstraction to a multiplicative interaction to improve the accuracy, as backpropagation found solutions that favors relying on the visual input features rather than image class input (Table S\ref{tab:cifar100-methods-results_concatenated_latent}). This observation coupled with the overall low accuracy from the pure generative model suggests that image class input does not significantly reduce the variance seen in pixel space, leading the decoder to favor the visual features from the encoder. We anticipate that scaling to larger datasets will benefit from richer task abstractions. This highlights a role for algorithms that discover such abstractions from data \citep{butz_learning_2019, hummos_thalamus_2023}.

\begin{wraptable}{l}{0.5\textwidth}
    \centering
    \caption{GBI accuracy on CIFAR100 using a multiplicative task abstraction. The task abstraction is projected through a frozen embedding layer (~Bernoulli distribution) then multiplies the visual information going to decoder.}
    \label{table:cifar100-methods-results}
\resizebox{.5\textwidth}{!}{
    \begin{tabular}{lcc}
        \hline
        \textbf{Method} & \textbf{Test Accuracy (\%)} & \textbf{Model runs}\\
        \hline
        Pure generative & 9.57 $\pm$ 0.19 & 100\\
        \hline
        GBI & 18.52 $\pm$ 0.38 & 1\\
        Iterative optimization & 18.53 $\pm$ 0.38 & 400\\
        Likelihood & 21.30 $\pm$ 0.36 & 100\\
        \hline
    \end{tabular}
    }
\end{wraptable}

\subsection{Experiment 3: Task-aware language modeling}
\label{section:language}
We apply the concepts demonstrated in the Bayesian toy dataset and the MNIST classifier to language modelling for a self-supervised sequence modeling task. Specifically, we are interested in the effects of providing a task representation on data-efficiency, training dynamics and the ability to generalize, as observed in the case of the Bayesian toy dataset (Fig \ref{fig:toy_task}, \ref{fig:toy_task_training_losses}, \ref{fig:toy_task_GBI_results}D,E). 

\paragraph*{Method} We train an LSTM (with 2 layers, 200 units each) on word-level language prediction and compare an LSTM with our GBI-LSTM. To demonstrate the effectiveness of task abstractions in this setting, we represent each task abstraction as a one-hot encoded identifier of each training dataset. We use datasets included in the BabyLM challenge \citep{warstadt_call_2023}, which provides 10 datasets inspired by what language data children might be exposed to during development, including wikipedia dataset, movie subtitles, children stories, and adult conversations, amongst others (listed in Appendix E). We train the two models on 3 of these datasets in a setting analogous to the Bayesian model dataset described earlier with data in each batch randomly drawn from one of the three datasets (we vary the choice of the 3 training datasets across seeds). The GBI-LSTM receives additional input with a one-hot encoded dataset identifier during training concatenated to the input token at each time step.

During testing, we use one-step gradients to infer the dataset from the text data provided as we assume no access to the dataset identifier at test time. We then use iterative optimization of the task abstraction input z to adapt the model to novel datasets by taking gradients steps on one batch of data and evaluating on the test set. 

\begin{figure}[t]
    \centering
    \includegraphics[width=1.\textwidth]{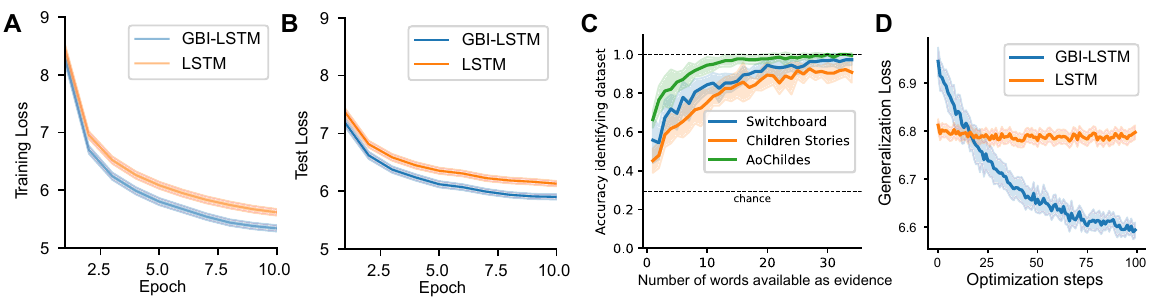}
    \caption{Better data-efficiency in an LSTM trained with task abstraction input (dataset ID one-hot) (GBI-LSTM) compared to training without such input (LSTM). Cross entropy loss for GBI-LSTM decreases faster for GBI-LSTM on the A) training set and B) testing set. Note that we train with a word-level tokenizer with a larger output dimension which has a highe cross entropy loss upper bound (~12). Each simulation run had three language datasets picked randomly from 10 datasets. 10 data set picks, each run for 3 random seeds, shaded area SEM. C) One-pass gradient-based inference can infer dataset ID accurately at test time needing only a few words. GBI-LSTM takes a sequence of words available and outputs predictions over the same sequence shifted by one token (sequence to sequence modeling). We take the gradients w.r.t the task abstraction inputs and use the gradients to infer dataset ID. We show examples from 3 datasets to show variability. D) Iterative optimization can be used to adapt a GBI-LSTM to a novel dataset. We compare it to optimizing the inputs of an LSTM that was trained with no task provided to rule out any generic optimization effects. Due to the variability of loss values across datasets, for each dataset, we take the mean and SEM over 4 seeds and then aggregate the results across datasets.
    }
    \label{fig:language_losses}
\end{figure}

\paragraph*{Results} First, we focus on the effect of task abstractions on data-efficiency. Compared with the LSTM, GBI-LSTM displays lower training and validation losses (Fig \ref{fig:language_losses}A, B). While the improvement is modest, we emphasize that we only had to add \textless $1.6$ bits of information to obtain it.

    

Second, we focus on task inference. We show that GBI can quickly and reliably infer the dataset at hand, and in fact, it takes a few words from the first batch to accurately infer the dataset (Fig \ref{fig:language_losses}C). Note that similar to likelihood functions, gradient inference can handle variable amounts of evidence, and the estimation accuracy scales with the amount of evidence available. 

Third, we focus on generalization. Our results show that task abstraction layers improves the model's ability to generalize to novel tasks (Fig \ref{fig:language_losses}D), similar to the case in the synthetic Bayesian dataset (Fig \ref{fig:toy_task_GBI_results}D,E). We use a 4th dataset not seen during training, and we compare the GBI-LSTM and LSTM losses on this novel dataset. To allow gradient descent to compose the closest solution to the novel datasets, we optimize the task abstraction input nodes of the GBI-LSTM to lower its prediction loss and we see a decrease in loss as it adjusts to this new unseen dataset (Fig \ref{fig:language_losses}D) (See Appendix E for example run). Notably, the GBI-LSTM starts out at a higher loss, on average, highlighting that task-aware models require  task inference. 

\section{Conclusions and future directions}
Overall we show that neural networks with access to task abstractions learn faster, forget less, adapt to changes in data distribution, and generalize to new situations. At the same time, we also provide tools for the networks to function independently infer their own task abstractions during testing. These task abstractions can be provided by humans, a trained teacher model, or inferred from data with no supervision (e.g. \citep{hummos_thalamus_2023}). Task-dependent networks require task inference and inferring an incorrect task impairs function. In animal brains, some computations that are frequently needed or are critical for survival might be better mapped to the default state of the brain. The brain shows distinct dynamics when not engaged in a specific task, the default mode, and we show previously unexplored connections between feedback signals in this default state and distributional inference signals that can support many probabilistic quantities such as uncertainty and surprise. We see future work addressing the limitation of relying on human-provided task abstractions as opposed to unsupervised discovery of richer multidimensional task representations along dimensions important for adaptation and generalization. Finally, while our method is theoretically motivated, we see future work extending these findings to larger datasets and more real-world applications.

\bibliography{main_z}

\begin{thebibliography}{59}
\providecommand{\natexlab}[1]{#1}
\providecommand{\url}[1]{\texttt{#1}}
\expandafter\ifx\csname urlstyle\endcsname\relax
  \providecommand{\doi}[1]{doi: #1}\else
  \providecommand{\doi}{doi: \begingroup \urlstyle{rm}\Url}\fi

\bibitem[Abdar et~al.(2021)Abdar, Pourpanah, Hussain, Rezazadegan, Liu, Ghavamzadeh, Fieguth, Cao, Khosravi, Acharya, Makarenkov, and Nahavandi]{abdar_review_2021}
Abdar, M., Pourpanah, F., Hussain, S., Rezazadegan, D., Liu, L., Ghavamzadeh, M., Fieguth, P., Cao, X., Khosravi, A., Acharya, U.~R., Makarenkov, V., and Nahavandi, S.
\newblock A review of uncertainty quantification in deep learning: {Techniques}, applications and challenges.
\newblock \emph{Information Fusion}, 76:\penalty0 243--297, December 2021.
\newblock ISSN 1566-2535.
\newblock \doi{10.1016/j.inffus.2021.05.008}.
\newblock URL \url{https://www.sciencedirect.com/science/article/pii/S1566253521001081}.

\bibitem[Abdelali et~al.(2014)Abdelali, Guzman, Sajjad, and Vogel]{abdelali_amara_2014}
Abdelali, A., Guzman, F., Sajjad, H., and Vogel, S.
\newblock The {AMARA} {Corpus}: {Building} {Parallel} {Language} {Resources} for the {Educational} {Domain}.
\newblock January 2014.
\newblock \doi{10.13140/2.1.1806.2406}.

\bibitem[Andreas et~al.(2016)Andreas, Rohrbach, Darrell, and Klein]{andreas_neural_2016}
Andreas, J., Rohrbach, M., Darrell, T., and Klein, D.
\newblock Neural module networks.
\newblock In \emph{Proceedings of the {IEEE} conference on computer vision and pattern recognition}, pp.\  39--48, 2016.

\bibitem[Badre et~al.(2010)Badre, Kayser, and D'Esposito]{badre_frontal_2010}
Badre, D., Kayser, A.~S., and D'Esposito, M.
\newblock Frontal cortex and the discovery of abstract action rules.
\newblock \emph{Neuron}, 66\penalty0 (2):\penalty0 315--326, April 2010.
\newblock ISSN 1097-4199.
\newblock \doi{10.1016/j.neuron.2010.03.025}.

\bibitem[Blundell et~al.(2015)Blundell, Cornebise, Kavukcuoglu, and Wierstra]{blundell_weight_2015}
Blundell, C., Cornebise, J., Kavukcuoglu, K., and Wierstra, D.
\newblock Weight {Uncertainty} in {Neural} {Network}.
\newblock In \emph{Proceedings of the 32nd {International} {Conference} on {Machine} {Learning}}, pp.\  1613--1622. PMLR, June 2015.
\newblock URL \url{https://proceedings.mlr.press/v37/blundell15.html}.
\newblock ISSN: 1938-7228.

\bibitem[Butz et~al.(2019)Butz, Bilkey, Humaidan, Knott, and Otte]{butz_learning_2019}
Butz, M.~V., Bilkey, D., Humaidan, D., Knott, A., and Otte, S.
\newblock Learning, planning, and control in a monolithic neural event inference architecture.
\newblock \emph{Neural Networks}, 117:\penalty0 135--144, September 2019.
\newblock ISSN 0893-6080.
\newblock \doi{10.1016/j.neunet.2019.05.001}.
\newblock URL \url{https://www.sciencedirect.com/science/article/pii/S0893608019301339}.

\bibitem[Castañón et~al.(2021)Castañón, Cardoso-Leite, Altarelli, Green, Schrater, and Bavelier]{castanon_mixture_2021}
Castañón, S.~H., Cardoso-Leite, P., Altarelli, I., Green, C.~S., Schrater, P., and Bavelier, D.
\newblock A mixture of generative models strategy helps humans generalize across tasks.
\newblock \emph{bioRxiv}, pp.\  2021.02.16.431506, January 2021.
\newblock \doi{10.1101/2021.02.16.431506}.
\newblock URL \url{http://biorxiv.org/content/early/2021/02/16/2021.02.16.431506.abstract}.

\bibitem[Collins \& Koechlin(2012)Collins and Koechlin]{collins_reasoning_2012}
Collins, A. and Koechlin, E.
\newblock Reasoning, {Learning}, and {Creativity}: {Frontal} {Lobe} {Function} and {Human} {Decision}-{Making}.
\newblock \emph{PLOS Biology}, 10\penalty0 (3):\penalty0 e1001293, March 2012.
\newblock ISSN 1545-7885.
\newblock \doi{10.1371/journal.pbio.1001293}.
\newblock URL \url{https://journals.plos.org/plosbiology/article?id=10.1371/journal.pbio.1001293}.
\newblock Publisher: Public Library of Science.

\bibitem[Collins(2017)]{collins_cost_2017}
Collins, A. G.~E.
\newblock The {Cost} of {Structure} {Learning}.
\newblock \emph{Journal of Cognitive Neuroscience}, 29\penalty0 (10):\penalty0 1646--1655, October 2017.
\newblock ISSN 0898-929X.
\newblock \doi{10.1162/jocn_a_01128}.
\newblock URL \url{https://doi.org/10.1162/jocn\_a\_01128}.
\newblock \_eprint: https://direct.mit.edu/jocn/article-pdf/29/10/1646/1953060/jocn\_a\_01128.pdf.

\bibitem[Fetaya et~al.(2020)Fetaya, Jacobsen, Grathwohl, and Zemel]{fetaya_understanding_2020}
Fetaya, E., Jacobsen, J.-H., Grathwohl, W., and Zemel, R.
\newblock Understanding the {Limitations} of {Conditional} {Generative} {Models}.
\newblock In \emph{International {Conference} on {Learning} {Representations}}, 2020.
\newblock URL \url{https://openreview.net/forum?id=r1lPleBFvH}.

\bibitem[Gaissmaier \& Schooler(2008)Gaissmaier and Schooler]{gaissmaier_smart_2008}
Gaissmaier, W. and Schooler, L.~J.
\newblock The smart potential behind probability matching.
\newblock \emph{Cognition}, 109\penalty0 (3):\penalty0 416--422, December 2008.
\newblock ISSN 0010-0277.
\newblock \doi{10.1016/j.cognition.2008.09.007}.
\newblock URL \url{https://www.sciencedirect.com/science/article/pii/S0010027708002151}.

\bibitem[Gal \& Ghahramani(2016)Gal and Ghahramani]{gal_dropout_2016}
Gal, Y. and Ghahramani, Z.
\newblock Dropout as a {Bayesian} {Approximation}: {Representing} {Model} {Uncertainty} in {Deep} {Learning}.
\newblock In \emph{Proceedings of {The} 33rd {International} {Conference} on {Machine} {Learning}}, pp.\  1050--1059. PMLR, June 2016.
\newblock URL \url{https://proceedings.mlr.press/v48/gal16.html}.
\newblock ISSN: 1938-7228.

\bibitem[Godfrey et~al.(1992)Godfrey, Holliman, and McDaniel]{godfrey_switchboard_1992}
Godfrey, J., Holliman, E., and McDaniel, J.
\newblock {SWITCHBOARD}: telephone speech corpus for research and development.
\newblock In \emph{[{Proceedings}] {ICASSP}-92: 1992 {IEEE} {International} {Conference} on {Acoustics}, {Speech}, and {Signal} {Processing}}, volume~1, pp.\  517--520 vol.1, March 1992.
\newblock \doi{10.1109/ICASSP.1992.225858}.
\newblock ISSN: 1520-6149.

\bibitem[Goyal et~al.(2019)Goyal, Lamb, Hoffmann, Sodhani, Levine, Bengio, and Schölkopf]{goyal_recurrent_2019}
Goyal, A., Lamb, A., Hoffmann, J., Sodhani, S., Levine, S., Bengio, Y., and Schölkopf, B.
\newblock Recurrent {Independent} {Mechanisms}.
\newblock \emph{ArXiv}, 2019.

\bibitem[Graves(2011)]{graves_practical_2011}
Graves, A.
\newblock Practical {Variational} {Inference} for {Neural} {Networks}.
\newblock In \emph{Advances in {Neural} {Information} {Processing} {Systems}}, volume~24. Curran Associates, Inc., 2011.
\newblock URL \url{https://papers.nips.cc/paper_files/paper/2011/hash/7eb3c8be3d411e8ebfab08eba5f49632-Abstract.html}.

\bibitem[Guo et~al.(2017)Guo, Pleiss, Sun, and Weinberger]{guo_calibration_2017}
Guo, C., Pleiss, G., Sun, Y., and Weinberger, K.~Q.
\newblock On {Calibration} of {Modern} {Neural} {Networks}.
\newblock In \emph{Proceedings of the 34th {International} {Conference} on {Machine} {Learning}}, pp.\  1321--1330. PMLR, July 2017.
\newblock URL \url{https://proceedings.mlr.press/v70/guo17a.html}.
\newblock ISSN: 2640-3498.

\bibitem[Hendrycks \& Gimpel(2018)Hendrycks and Gimpel]{hendrycks_baseline_2018}
Hendrycks, D. and Gimpel, K.
\newblock A {Baseline} for {Detecting} {Misclassified} and {Out}-of-{Distribution} {Examples} in {Neural} {Networks}, October 2018.
\newblock URL \url{http://arxiv.org/abs/1610.02136}.
\newblock arXiv:1610.02136 [cs].

\bibitem[Hill et~al.(2016)Hill, Bordes, Chopra, and Weston]{hill_goldilocks_2016}
Hill, F., Bordes, A., Chopra, S., and Weston, J.
\newblock The {Goldilocks} {Principle}: {Reading} {Children}'s {Books} with {Explicit} {Memory} {Representations}, April 2016.
\newblock URL \url{http://arxiv.org/abs/1511.02301}.
\newblock arXiv:1511.02301 [cs].

\bibitem[Hochreiter \& Schmidhuber(1997)Hochreiter and Schmidhuber]{hochreiter_long_1997}
Hochreiter, S. and Schmidhuber, J.
\newblock Long {Short}-{Term} {Memory}.
\newblock \emph{Neural Computation}, 9\penalty0 (8):\penalty0 1735--1780, November 1997.
\newblock ISSN 0899-7667, 1530-888X.
\newblock \doi{10.1162/neco.1997.9.8.1735}.
\newblock URL \url{https://direct.mit.edu/neco/article/9/8/1735-1780/6109}.

\bibitem[Houlsby et~al.(2019)Houlsby, Giurgiu, Jastrzebski, Morrone, Laroussilhe, Gesmundo, Attariyan, and Gelly]{houlsby_parameter-efficient_2019}
Houlsby, N., Giurgiu, A., Jastrzebski, S., Morrone, B., Laroussilhe, Q.~D., Gesmundo, A., Attariyan, M., and Gelly, S.
\newblock Parameter-{Efficient} {Transfer} {Learning} for {NLP}.
\newblock In \emph{Proceedings of the 36th {International} {Conference} on {Machine} {Learning}}, pp.\  2790--2799. PMLR, May 2019.
\newblock URL \url{https://proceedings.mlr.press/v97/houlsby19a.html}.
\newblock ISSN: 2640-3498.

\bibitem[Hu et~al.(2021)Hu, Shen, Wallis, Allen-Zhu, Li, Wang, Wang, and Chen]{hu_lora_2021}
Hu, E.~J., Shen, Y., Wallis, P., Allen-Zhu, Z., Li, Y., Wang, S., Wang, L., and Chen, W.
\newblock {LoRA}: {Low}-{Rank} {Adaptation} of {Large} {Language} {Models}, October 2021.
\newblock URL \url{http://arxiv.org/abs/2106.09685}.
\newblock arXiv:2106.09685 [cs].

\bibitem[Hummos(2023)]{hummos_thalamus_2023}
Hummos, A.
\newblock Thalamus: a brain-inspired algorithm for biologically-plausible continual learning and disentangled representations.
\newblock In \emph{The {Eleventh} {International} {Conference} on {Learning} {Representations}}, 2023.
\newblock URL \url{https://openreview.net/forum?id=6orC5MvgPBK}.

\bibitem[Hummos et~al.(2022)Hummos, Wang, Drammis, Halassa, and Pleger]{hummos_thalamic_2022}
Hummos, A., Wang, B.~A., Drammis, S., Halassa, M.~M., and Pleger, B.
\newblock Thalamic regulation of frontal interactions in human cognitive flexibility.
\newblock \emph{PLOS Computational Biology}, 18\penalty0 (9):\penalty0 e1010500, September 2022.
\newblock ISSN 1553-7358.
\newblock \doi{10.1371/journal.pcbi.1010500}.
\newblock URL \url{https://journals.plos.org/ploscompbiol/article?id=10.1371/journal.pcbi.1010500}.
\newblock Publisher: Public Library of Science.

\bibitem[Hurtado et~al.(2021)Hurtado, Raymond, and Soto]{hurtado_optimizing_2021}
Hurtado, J., Raymond, A., and Soto, A.
\newblock Optimizing {Reusable} {Knowledge} for {Continual} {Learning} via {Metalearning}.
\newblock In \emph{Advances in {Neural} {Information} {Processing} {Systems}}, volume~34, pp.\  14150--14162. Curran Associates, Inc., 2021.
\newblock URL \url{https://proceedings.neurips.cc/paper/2021/hash/761e6675f9e54673cc778e7fdb2823d2-Abstract.html}.

\bibitem[Jordan et~al.(1998)Jordan, Ghahramani, Jaakkola, and Saul]{jordan_introduction_1998}
Jordan, M.~I., Ghahramani, Z., Jaakkola, T.~S., and Saul, L.~K.
\newblock An {Introduction} to {Variational} {Methods} for {Graphical} {Models}.
\newblock In Jordan, M.~I. (ed.), \emph{Learning in {Graphical} {Models}}, pp.\  105--161. Springer Netherlands, Dordrecht, 1998.
\newblock ISBN 978-94-010-6104-9 978-94-011-5014-9.
\newblock \doi{10.1007/978-94-011-5014-9_5}.
\newblock URL \url{http://link.springer.com/10.1007/978-94-011-5014-9_5}.

\bibitem[Kingma \& Welling(2013)Kingma and Welling]{kingma_auto-encoding_2013}
Kingma, D.~P. and Welling, M.
\newblock Auto-{Encoding} {Variational} {Bayes}.
\newblock 2013.
\newblock \doi{10.48550/ARXIV.1312.6114}.
\newblock URL \url{https://arxiv.org/abs/1312.6114}.
\newblock Publisher: arXiv Version Number: 11.

\bibitem[Kirkpatrick et~al.(2017)Kirkpatrick, Pascanu, Rabinowitz, Veness, Desjardins, Rusu, Milan, Quan, Ramalho, Grabska-Barwinska, Hassabis, Clopath, Kumaran, and Hadsell]{kirkpatrick_overcoming_2017}
Kirkpatrick, J., Pascanu, R., Rabinowitz, N., Veness, J., Desjardins, G., Rusu, A.~A., Milan, K., Quan, J., Ramalho, T., Grabska-Barwinska, A., Hassabis, D., Clopath, C., Kumaran, D., and Hadsell, R.
\newblock Overcoming catastrophic forgetting in neural networks.
\newblock \emph{Proceedings of the National Academy of Sciences}, 114\penalty0 (13):\penalty0 3521--3526, March 2017.
\newblock \doi{10.1073/pnas.1611835114}.
\newblock URL \url{https://www.pnas.org/doi/10.1073/pnas.1611835114}.
\newblock Publisher: Proceedings of the National Academy of Sciences.

\bibitem[Kirsch et~al.(2018)Kirsch, Kunze, and Barber]{kirsch_modular_2018}
Kirsch, L., Kunze, J., and Barber, D.
\newblock Modular {Networks}: {Learning} to {Decompose} {Neural} {Computation}.
\newblock November 2018.
\newblock URL \url{https://www.semanticscholar.org/paper/Modular-Networks%3A-Learning-to-Decompose-Neural-Kirsch-Kunze/0b50b9e103e19d87c2d30ed0d157d8379320ce6f}.

\bibitem[Lahiri(2014)]{lahiri_complexity_2014}
Lahiri, S.
\newblock Complexity of {Word} {Collocation} {Networks}: {A} {Preliminary} {Structural} {Analysis}, March 2014.
\newblock URL \url{http://arxiv.org/abs/1310.5111}.
\newblock arXiv:1310.5111 [physics] version: 2.

\bibitem[Lake et~al.(2017)Lake, Ullman, Tenenbaum, and Gershman]{lake_building_2017}
Lake, B.~M., Ullman, T.~D., Tenenbaum, J.~B., and Gershman, S.~J.
\newblock Building machines that learn and think like people.
\newblock \emph{Behavioral and Brain Sciences}, 40:\penalty0 e253, 2017.
\newblock ISSN 0140-525X, 1469-1825.
\newblock \doi{10.1017/S0140525X16001837}.
\newblock URL \url{https://www.cambridge.org/core/product/identifier/S0140525X16001837/type/journal_article}.

\bibitem[Lakshminarayanan et~al.(2017)Lakshminarayanan, Pritzel, and Blundell]{lakshminarayanan_simple_2017}
Lakshminarayanan, B., Pritzel, A., and Blundell, C.
\newblock Simple and {Scalable} {Predictive} {Uncertainty} {Estimation} using {Deep} {Ensembles}.
\newblock In \emph{Advances in {Neural} {Information} {Processing} {Systems}}, volume~30. Curran Associates, Inc., 2017.
\newblock URL \url{https://proceedings.neurips.cc/paper/2017/hash/9ef2ed4b7fd2c810847ffa5fa85bce38-Abstract.html}.

\bibitem[Lester et~al.(2021)Lester, Al-Rfou, and Constant]{lester_power_2021}
Lester, B., Al-Rfou, R., and Constant, N.
\newblock The {Power} of {Scale} for {Parameter}-{Efficient} {Prompt} {Tuning}.
\newblock Technical Report arXiv:2104.08691, arXiv, September 2021.
\newblock URL \url{http://arxiv.org/abs/2104.08691}.
\newblock arXiv:2104.08691 [cs] type: article.

\bibitem[Li \& Liang(2021)Li and Liang]{li_prefix-tuning_2021}
Li, X.~L. and Liang, P.
\newblock Prefix-{Tuning}: {Optimizing} {Continuous} {Prompts} for {Generation}.
\newblock In \emph{Proceedings of the 59th {Annual} {Meeting} of the {Association} for {Computational} {Linguistics} and the 11th {International} {Joint} {Conference} on {Natural} {Language} {Processing} ({Volume} 1: {Long} {Papers})}, pp.\  4582--4597, Online, August 2021. Association for Computational Linguistics.
\newblock \doi{10.18653/v1/2021.acl-long.353}.
\newblock URL \url{https://aclanthology.org/2021.acl-long.353}.

\bibitem[Lison \& Tiedemann(2016)Lison and Tiedemann]{lison_opensubtitles2016_2016}
Lison, P. and Tiedemann, J.
\newblock {OpenSubtitles2016}: {Extracting} {Large} {Parallel} {Corpora} from {Movie} and {TV} {Subtitles}.
\newblock In \emph{Proceedings of the {Tenth} {International} {Conference} on {Language} {Resources} and {Evaluation} ({LREC}'16)}, pp.\  923--929, Portorož, Slovenia, May 2016. European Language Resources Association (ELRA).
\newblock URL \url{https://aclanthology.org/L16-1147}.

\bibitem[MacKay(1992)]{mackay_practical_1992}
MacKay, D. J.~C.
\newblock A {Practical} {Bayesian} {Framework} for {Backpropagation} {Networks}.
\newblock \emph{Neural Computation}, 4\penalty0 (3):\penalty0 448--472, May 1992.
\newblock ISSN 0899-7667, 1530-888X.
\newblock \doi{10.1162/neco.1992.4.3.448}.
\newblock URL \url{https://direct.mit.edu/neco/article/4/3/448-472/5654}.

\bibitem[MacWhinney(2000)]{macwhinney_childes_2000}
MacWhinney, B.
\newblock \emph{The {CHILDES} project: {Tools} for analyzing talk: {Transcription} format and programs, {Vol}. 1, 3rd ed.}
\newblock The {CHILDES} project: {Tools} for analyzing talk: {Transcription} format and programs, {Vol}. 1, 3rd ed. Lawrence Erlbaum Associates Publishers, Mahwah, NJ, US, 2000.
\newblock ISBN 0-8058-2995-4 (Hardcover).
\newblock Pages: xi, 366.

\bibitem[Mante et~al.(2013)Mante, Sussillo, Shenoy, and Newsome]{mante_context-dependent_2013}
Mante, V., Sussillo, D., Shenoy, K.~V., and Newsome, W.~T.
\newblock Context-dependent computation by recurrent dynamics in prefrontal cortex.
\newblock \emph{Nature}, 503\penalty0 (7474):\penalty0 78--84, November 2013.
\newblock ISSN 1476-4687.
\newblock \doi{10.1038/nature12742}.
\newblock URL \url{https://www.nature.com/articles/nature12742}.
\newblock Publisher: Nature Publishing Group.

\bibitem[Marino et~al.(2018)Marino, Yue, and Mandt]{marino_iterative_2018}
Marino, J., Yue, Y., and Mandt, S.
\newblock Iterative {Amortized} {Inference}.
\newblock \emph{Proceedings of Machine Learning Research}, 80:\penalty0 3403--3412, July 2018.
\newblock ISSN 1938-7228.
\newblock URL \url{https://resolver.caltech.edu/CaltechAUTHORS:20190506-143507942}.
\newblock Conference Name: 35th International Conference on Machine Learning Meeting Name: 35th International Conference on Machine Learning Place: Stockholm, Sweden Publisher: PMLR.

\bibitem[Masse et~al.(2018)Masse, Grant, and Freedman]{masse_alleviating_2018}
Masse, N.~Y., Grant, G.~D., and Freedman, D.~J.
\newblock Alleviating catastrophic forgetting using context-dependent gating and synaptic stabilization.
\newblock \emph{Proceedings of the National Academy of Sciences}, 115\penalty0 (44):\penalty0 E10467--E10475, October 2018.
\newblock \doi{10.1073/pnas.1803839115}.
\newblock URL \url{https://www.pnas.org/doi/10.1073/pnas.1803839115}.
\newblock Publisher: Proceedings of the National Academy of Sciences.

\bibitem[Mazzaglia et~al.(2022)Mazzaglia, Verbelen, Dhoedt, Lacoste, and Rajeswar]{mazzaglia_choreographer_2022}
Mazzaglia, P., Verbelen, T., Dhoedt, B., Lacoste, A., and Rajeswar, S.
\newblock Choreographer: {Learning} and adapting skills in imagination.
\newblock \emph{arXiv preprint arXiv:2211.13350}, 2022.

\bibitem[Miller \& Cohen(2001)Miller and Cohen]{miller_integrative_2001}
Miller, E.~K. and Cohen, J.~D.
\newblock An integrative theory of prefrontal cortex function.
\newblock \emph{Annual Review of Neuroscience}, 24:\penalty0 167--202, 2001.
\newblock ISSN 0147-006X.
\newblock \doi{10.1146/annurev.neuro.24.1.167}.

\bibitem[Nalisnick et~al.(2019)Nalisnick, Matsukawa, Teh, Gorur, and Lakshminarayanan]{nalisnick_deep_2019}
Nalisnick, E., Matsukawa, A., Teh, Y.~W., Gorur, D., and Lakshminarayanan, B.
\newblock Do {Deep} {Generative} {Models} {Know} {What} {They} {Don}'t {Know}?
\newblock In \emph{International {Conference} on {Learning} {Representations}}, 2019.
\newblock URL \url{https://openreview.net/forum?id=H1xwNhCcYm}.

\bibitem[Neal \& Hinton(1998)Neal and Hinton]{neal_view_1998}
Neal, R.~M. and Hinton, G.~E.
\newblock A {View} of the {Em} {Algorithm} that {Justifies} {Incremental}, {Sparse}, and other {Variants}.
\newblock In Jordan, M.~I. (ed.), \emph{Learning in {Graphical} {Models}}, {NATO} {ASI} {Series}, pp.\  355--368. Springer Netherlands, Dordrecht, 1998.
\newblock ISBN 978-94-011-5014-9.
\newblock \doi{10.1007/978-94-011-5014-9_12}.
\newblock URL \url{https://doi.org/10.1007/978-94-011-5014-9_12}.

\bibitem[Niv(2019)]{niv_learning_2019}
Niv, Y.
\newblock Learning task-state representations.
\newblock \emph{Nature Neuroscience}, 22\penalty0 (10):\penalty0 1544--1553, October 2019.
\newblock ISSN 1546-1726.
\newblock \doi{10.1038/s41593-019-0470-8}.
\newblock URL \url{https://www.nature.com/articles/s41593-019-0470-8}.
\newblock Number: 10 Publisher: Nature Publishing Group.

\bibitem[Rezende et~al.(2014)Rezende, Mohamed, and Wierstra]{rezende_stochastic_2014}
Rezende, D.~J., Mohamed, S., and Wierstra, D.
\newblock Stochastic {Backpropagation} and {Approximate} {Inference} in {Deep} {Generative} {Models}.
\newblock In \emph{Proceedings of the 31st {International} {Conference} on {Machine} {Learning}}, pp.\  1278--1286. PMLR, June 2014.
\newblock URL \url{https://proceedings.mlr.press/v32/rezende14.html}.
\newblock ISSN: 1938-7228.

\bibitem[Rikhye et~al.(2018)Rikhye, Gilra, and Halassa]{rikhye_thalamic_2018}
Rikhye, R.~V., Gilra, A., and Halassa, M.~M.
\newblock Thalamic regulation of switching between cortical representations enables cognitive flexibility.
\newblock \emph{Nature Neuroscience}, 21\penalty0 (12):\penalty0 1753--1763, December 2018.
\newblock ISSN 1546-1726.
\newblock \doi{10.1038/s41593-018-0269-z}.

\bibitem[Roy et~al.(2024)Roy, Verma, and Gupta]{roy_efficient_2024}
Roy, S., Verma, V., and Gupta, D.
\newblock Efficient {Expansion} and {Gradient} {Based} {Task} {Inference} for {Replay} {Free} {Incremental} {Learning}.
\newblock In \emph{Proceedings of the {IEEE}/{CVF} {Winter} {Conference} on {Applications} of {Computer} {Vision} ({WACV})}, pp.\  1165--1175, January 2024.

\bibitem[Srivastava et~al.(2014)Srivastava, Hinton, Krizhevsky, Sutskever, and Salakhutdinov]{srivastava_dropout_2014}
Srivastava, N., Hinton, G., Krizhevsky, A., Sutskever, I., and Salakhutdinov, R.
\newblock Dropout: {A} {Simple} {Way} to {Prevent} {Neural} {Networks} from {Overfitting}.
\newblock \emph{Journal of Machine Learning Research}, 15\penalty0 (56):\penalty0 1929--1958, 2014.
\newblock ISSN 1533-7928.
\newblock URL \url{http://jmlr.org/papers/v15/srivastava14a.html}.

\bibitem[Tafazoli et~al.(2024)Tafazoli, Bouchacourt, Ardalan, Markov, Uchimura, Mattar, Daw, and Buschman]{tafazoli_building_2024}
Tafazoli, S., Bouchacourt, F.~M., Ardalan, A., Markov, N.~T., Uchimura, M., Mattar, M.~G., Daw, N.~D., and Buschman, T.~J.
\newblock Building compositional tasks with shared neural subspaces.
\newblock \emph{bioRxiv}, 2024.
\newblock \doi{10.1101/2024.01.31.578263}.
\newblock URL \url{https://www.biorxiv.org/content/early/2024/03/22/2024.01.31.578263}.
\newblock Publisher: Cold Spring Harbor Laboratory \_eprint: https://www.biorxiv.org/content/early/2024/03/22/2024.01.31.578263.full.pdf.

\bibitem[Vaidya et~al.(2021)Vaidya, Jones, Castillo, and Badre]{vaidya_neural_2021}
Vaidya, A.~R., Jones, H.~M., Castillo, J., and Badre, D.
\newblock Neural representation of abstract task structure during generalization.
\newblock \emph{eLife}, 10:\penalty0 e63226, March 2021.
\newblock ISSN 2050-084X.
\newblock \doi{10.7554/eLife.63226}.
\newblock URL \url{https://doi.org/10.7554/eLife.63226}.
\newblock Publisher: eLife Sciences Publications, Ltd.

\bibitem[Wang \& Zhang(2022)Wang and Zhang]{wang_contextual_2022}
Wang, D. and Zhang, S.
\newblock Contextual {Instance} {Decoupling} for {Robust} {Multi}-{Person} {Pose} {Estimation}.
\newblock pp.\  11060--11068, 2022.
\newblock URL \url{https://openaccess.thecvf.com/content/CVPR2022/html/Wang_Contextual_Instance_Decoupling_for_Robust_Multi-Person_Pose_Estimation_CVPR_2022_paper.html}.

\bibitem[Warstadt et~al.(2023)Warstadt, Choshen, Mueller, Williams, Wilcox, and Zhuang]{warstadt_call_2023}
Warstadt, A., Choshen, L., Mueller, A., Williams, A., Wilcox, E., and Zhuang, C.
\newblock Call for {Papers} -- {The} {BabyLM} {Challenge}: {Sample}-efficient pretraining on a developmentally plausible corpus, January 2023.
\newblock URL \url{https://arxiv.org/abs/2301.11796v1}.

\bibitem[Wortsman et~al.(2020)Wortsman, Ramanujan, Liu, Kembhavi, Rastegari, Yosinski, and Farhadi]{wortsman_supermasks_2020}
Wortsman, M., Ramanujan, V., Liu, R., Kembhavi, A., Rastegari, M., Yosinski, J., and Farhadi, A.
\newblock Supermasks in {Superposition}.
\newblock In Larochelle, H., Ranzato, M., Hadsell, R., Balcan, M.~F., and Lin, H. (eds.), \emph{Advances in {Neural} {Information} {Processing} {Systems}}, volume~33, pp.\  15173--15184. Curran Associates, Inc., 2020.
\newblock URL \url{https://proceedings.neurips.cc/paper_files/paper/2020/file/ad1f8bb9b51f023cdc80cf94bb615aa9-Paper.pdf}.

\bibitem[Xiao et~al.(2017)Xiao, Rasul, and Vollgraf]{xiao_fashion-mnist_2017}
Xiao, H., Rasul, K., and Vollgraf, R.
\newblock Fashion-{MNIST}: a {Novel} {Image} {Dataset} for {Benchmarking} {Machine} {Learning} {Algorithms}.
\newblock \emph{CoRR}, abs/1708.07747, 2017.
\newblock URL \url{http://arxiv.org/abs/1708.07747}.
\newblock \_eprint: 1708.07747.

\bibitem[Xiao et~al.(2020)Xiao, Yan, and Amit]{xiao_likelihood_2020}
Xiao, Z., Yan, Q., and Amit, Y.
\newblock Likelihood regret: {An} out-of-distribution detection score for variational auto-encoder.
\newblock \emph{Advances in neural information processing systems}, 33:\penalty0 20685--20696, 2020.

\bibitem[Xie et~al.(2021)Xie, Harrison, and Finn]{xie_deep_2021}
Xie, A., Harrison, J., and Finn, C.
\newblock Deep {Reinforcement} {Learning} amidst {Continual} {Structured} {Non}-{Stationarity}.
\newblock In Meila, M. and Zhang, T. (eds.), \emph{Proceedings of the 38th {International} {Conference} on {Machine} {Learning}}, volume 139 of \emph{Proceedings of {Machine} {Learning} {Research}}, pp.\  11393--11403. PMLR, July 2021.
\newblock URL \url{https://proceedings.mlr.press/v139/xie21c.html}.

\bibitem[Yang et~al.(2019)Yang, Joglekar, Song, Newsome, and Wang]{yang_task_2019}
Yang, G.~R., Joglekar, M.~R., Song, H.~F., Newsome, W.~T., and Wang, X.-J.
\newblock Task representations in neural networks trained to perform many cognitive tasks.
\newblock \emph{Nature Neuroscience}, 22\penalty0 (2):\penalty0 297--306, February 2019.
\newblock ISSN 1546-1726.
\newblock \doi{10.1038/s41593-018-0310-2}.
\newblock URL \url{https://www.nature.com/articles/s41593-018-0310-2}.

\bibitem[Yu et~al.(2021)Yu, Wilson, and Nassar]{yu_adaptive_2021}
Yu, L.~Q., Wilson, R.~C., and Nassar, M.~R.
\newblock Adaptive learning is structure learning in time.
\newblock \emph{Neuroscience \& Biobehavioral Reviews}, 128:\penalty0 270--281, September 2021.
\newblock ISSN 0149-7634.
\newblock \doi{10.1016/j.neubiorev.2021.06.024}.
\newblock URL \url{https://www.sciencedirect.com/science/article/pii/S0149763421002657}.

\bibitem[Zhou et~al.(2019)Zhou, Montesinos-Cartagena, Wikenheiser, Gardner, Niv, and Schoenbaum]{zhou_complementary_2019}
Zhou, J., Montesinos-Cartagena, M., Wikenheiser, A.~M., Gardner, M. P.~H., Niv, Y., and Schoenbaum, G.
\newblock Complementary {Task} {Structure} {Representations} in {Hippocampus} and {Orbitofrontal} {Cortex} during an {Odor} {Sequence} {Task}.
\newblock \emph{Current Biology}, 29\penalty0 (20):\penalty0 3402--3409.e3, October 2019.
\newblock ISSN 0960-9822.
\newblock \doi{10.1016/j.cub.2019.08.040}.
\newblock URL \url{https://www.sciencedirect.com/science/article/pii/S0960982219310905}.

\end{thebibliography}
\bibliographystyle{icml2024}

\newpage
\appendix
\onecolumn
\newpage

\section{Related Work}
\label{appendix A}


\subsection{Adapting large models}
Several methods have been developed for adapting large models to specific tasks without expensive full fine-tuning. Prompt and prefix tuning involve the use of learnable prompts \citep{lester_power_2021} or a continuous prefix \citep{li_prefix-tuning_2021} that guide the model's outputs. Lower-rank adaptation \citep{hu_lora_2021} modifies model parameters with a low-rank tensor, learned during adapting the model, while Adapters introduce small, task-specific modules within the network that are trained while the rest of the model remains fixed \citep{houlsby_parameter-efficient_2019}. Different from these methods, our approach to contextualizing a generative model proposes providing context signals or inferring them during the pretraining of the model. This allows the model to organize knowledge along these dimensions giving it an inductive bias anticipating variations along the same dimensions. 

\subsection{Iterative Variational Inference}
Variational inference, a method that recasts inference as an optimization problem \citep{jordan_introduction_1998, neal_view_1998}, has been instrumental to the development of deep variational models. However, recent work has primarily used the variational formulation for training, relying on fast amortized directed encoders to infer an approximate posterior distribution \citep{kingma_auto-encoding_2013, rezende_stochastic_2014}, thereby avoiding the potentially lengthy optimization loops of taking gradients through the decoder to refine the posterior distribution. To mitigate this, amortized variational inference employs an additional neural network that takes either the gradients from the decoder, or the reconstruction errors and updates the approximate posterior distribution with only a few iterations needed \citep{marino_iterative_2018}. Our work builds upon these insights, demonstrating that the initial gradient to the latent carries sufficient information to approximate likelihood functions. This reduces the need for multiple model runs or learning an additional model, but keeps iterative refinement as an option for complex samples.

\subsection{Quantifying Uncertainty}
Uncertainty quantification is a critical aspect of both Bayesian and neural network paradigms, with a rich and extensive history of literature, for review see \citep{abdar_review_2021}. Many approaches derive from the maximum likelihood interpretation of neural network training \citep{mackay_practical_1992}, known as Bayesian Neural Networks (BNNs). Such models model uncertainty by placing prior distributions over the model's weights, with Variational Inference and dropout-based methods (like MCdropout and MCdropConnect) serving as practical approximations \citep{blundell_weight_2015, graves_practical_2011, gal_dropout_2016, srivastava_dropout_2014}. Ensemble methods, such as Deep Ensembles, improve predictive performance and provide calibrated uncertainty measures by aggregating predictions of multiple independently trained neural networks \citep{lakshminarayanan_simple_2017}. Lastly, temperature scaling is a post-processing method used to calibrate the model's softmax outputs, aligning the model's confidence with its prediction accuracy \citep{guo_calibration_2017}.

\subsection{Out-of-Distribution (OOD) Detection}
The role of out-of-distribution (OOD) detection is essential for the secure implementation of machine learning systems. This is due to the tendency of models to deliver erroneously confident predictions when faced with data that diverges from the distribution on which they were initially trained. Current OOD detection methods fall primarily into two classes: discriminator-based either the logit or the feature space \citep{hendrycks_baseline_2018} or generation-based approaches which employ either the disparity in data reconstruction or the estimation of density in latent space \citep{nalisnick_deep_2019}, with the appealing intuition that generative models capture the data distribution and can detect out of distribution samples. While one might suspect that our method works because it adds a generative objective, recent work showed that several families of generative models might assign a higher probability to out-of-distribution data than in-distribution (\citep{nalisnick_deep_2019, fetaya_understanding_2020}). Using their rich visual latent spaces, they can generalize easily to other datasets. Different that these models, our model separates computation of visual features into one stream and the task abstractions into another, instead of assessing OOD probability from the visual features, we selectively assess in the task abstraction space.

\section{Bayesian Toy Dataset}
\label{appendix B}
\subsection{Bayesian graphical model task}

\subsubsection{Bayesian generative model}

The task involves a sequence of observations $x_t$ with a new observation at each time step $t$. The observations are generated from the following Bayesian model with context nodes z, a categorical variable, here 2-way (binary). Variable z is represented as one-hot encoded vector that selects the mean between two different configurations, while variance remains constant, as follows:

\begin{equation} 
    \vve{z}_0 = \text{random choice:  [0,1] or [1,0] }
\end{equation}
with a transition probability matrix:
\begin{equation} 
    P(\vve{z}_t | \vve{z}_{t-1}) = 
    \begin{bmatrix} 
	1-p_v & p_v \\
	p_v & 1-p_v\\
\end{bmatrix}
\end{equation}
Where $P_v$ (=0.05) is the volatility of the context, or the probability of context switching every time step. We enforce a minimum of 20 trials per block and a maximum of 50. 
\begin{equation} 
    \nonumber P(x_t| \vve{z}_t = [0,1]) = \mathcal{N}(0.8, \sigma)
\end{equation}
\begin{equation} 
    P(x_t| \vve{z}_t = [1,0]) = \mathcal{N}(0.2, \sigma)
\end{equation}

Then using Bayes rule we express the posterior over context units at newest time step $\vve{z}_{t+1}$ given the history of the observations as follows:
\begin{equation}
    P(z_{t+1}|x_{0:t+1}) = \frac{P(x_{t+1}|z_{t+1})P(z_{t+1}|z_{0:t}, x_{0:t})}{P(x_{t+1}|z_{0:t}, x_{0:t})}
    \label{eq:Bayes_dynamic}
\end{equation}
and the likelihood function as the Gaussian probability distribution density:
\begin{equation}
\label{eqn:gaussian_pdf}
P(x_t\mid z_t) = \frac{1}{\sqrt{2\pi \sigma^{2}}} \exp\left(-\frac{(x_{t} -\mu_{1})^2}{2\sigma^{2}}\right)
\end{equation}

\subsubsection{Neural Model Architecture and Training}

We employed the LSTM implementation provided by Pytorch, utilizing an input layer of size [1, 100] for mapping inputs to the LSTM, which consists of 100 hidden units. In the case of the GBI-LSTM the model received additional inputs from two additional units, the task abstraction units, after passing through a softmax activation function. An output layer of size [100, 2] was used to map to the output's mean and variance estimates. The model was trained to maximize the log-likelihood of the observed data by predicting a mean and a variance estimate, allowing the model to express a distribution over the next observation.. The final form of the objective derived from the Gaussian PDF has the MSE divided by the output variance.
The LSTM was trained using the Adam optimizer with a learning rate of $10^{-3}$. To optimize the context representations, we employed Adam with a learning rate of $10^{-2}$ and a decay rate of $10^{-3}$. 




\begin{figure}
    \centering
    \includegraphics{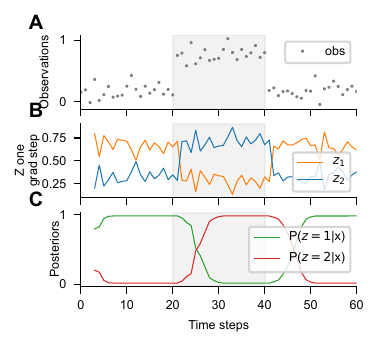}
    \caption{Grafting the neural gradients into Bayesian computations. With the task abstraction input z set to max entropy, we take one step along the gradients w.r.t to z. We use these one-step gradient updates in lieu of the likelihood function (eqn \ref{eqn:gaussian_pdf}), thereby avoiding having to discover the $\mu$ and $\sigma$ nodes of the underlying Bayesian causal model, and offloading the structure learning to the neural network, but maintaining the ability to proceed with Bayesian computations.}
    \label{fig:bayesian_gradients_to_bayesian}
\end{figure}

\begin{figure}
    \centering
    \includegraphics{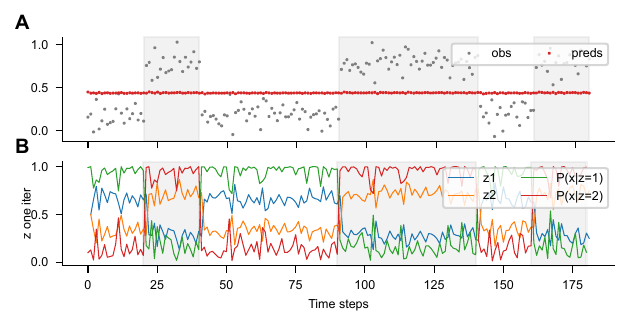}
    \caption{Overlaid likelihoods computed using the ground truth Bayesian model and neural gradients from GBI-LSTM for visual comparison. A) GBI-LSTM was in default mode with z at max entropy as responses and one step gradients recorded.}
    \label{fig:overlaid_gradients_and_likelihoods}
\end{figure}

\begin{figure}
    \centering
    \includegraphics{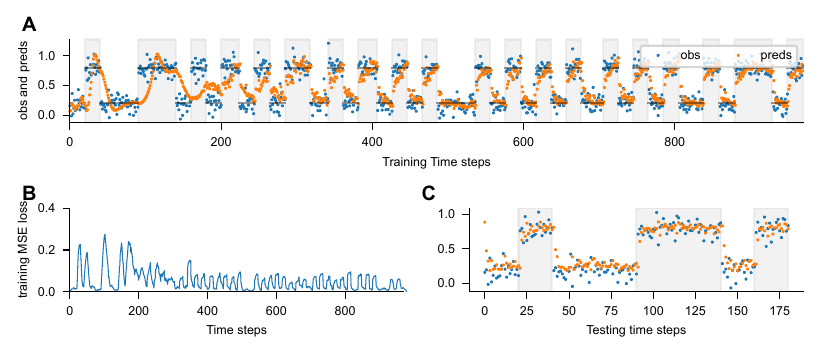}
    \caption{Training an LSTM on the Bayesian toy dataset, but no context from Bayesian nodes. An RNN trained on next observation prediction on this simple task learns it trivially well, and can adapt by detecting Gaussian distribution changes at context boundaries. Makes only one mistake typically before switching its predictions.}
    \label{fig:bayesian_lstm_no_context}
\end{figure}

\begin{figure}
    \centering
    \includegraphics{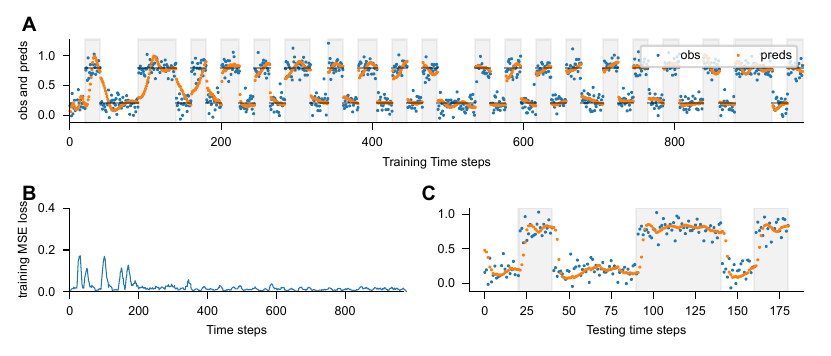}
    \caption{Training a GBI-LSTM on the Bayesian toy dataset with context from Bayesian model.}
    \label{fig:bayesian_lstm_with_context}
\end{figure}


\section{MNIST}
\label{appendix C}

\subsection{Further results on MNIST and confidence}

We compared the confidence values from the GBI autoencoder trained on MNIST digits to a convolutional classifier. The classifier and the encoder shared the same convolutional network, and the decoder used its transpose convolutions. Being interested in the errors the models make, we limited the capacity of the model and used a small number of convolutional filters, specifically we used these layers, (table \ref{table:covolutional_backbone}):

\begin{table}[]
    \centering
    \caption{Architecture of the convolutional classifier and encoder networks. Decoder had the same structure but transposed.}.
    \label{table:covolutional_backbone}
  \begin{tabular}{c|c|c|c|c}
\textbf{Layer} & \textbf{Input Channels} & \textbf{Output Channels} & \textbf{Stride} & \textbf{Activation} \\
\hline
Conv2d & $\text{image\_channels}$ & $2$ & $2$ & ReLU \\
Conv2d & $2$ & $4$ & $2$ & ReLU \\
Conv2d & $4$ & $8$ & - & None \\
Flatten & $8$ & - & - & - \\
\end{tabular}
\end{table}

\begin{figure}
  \centering
  \includegraphics[width=\textwidth]{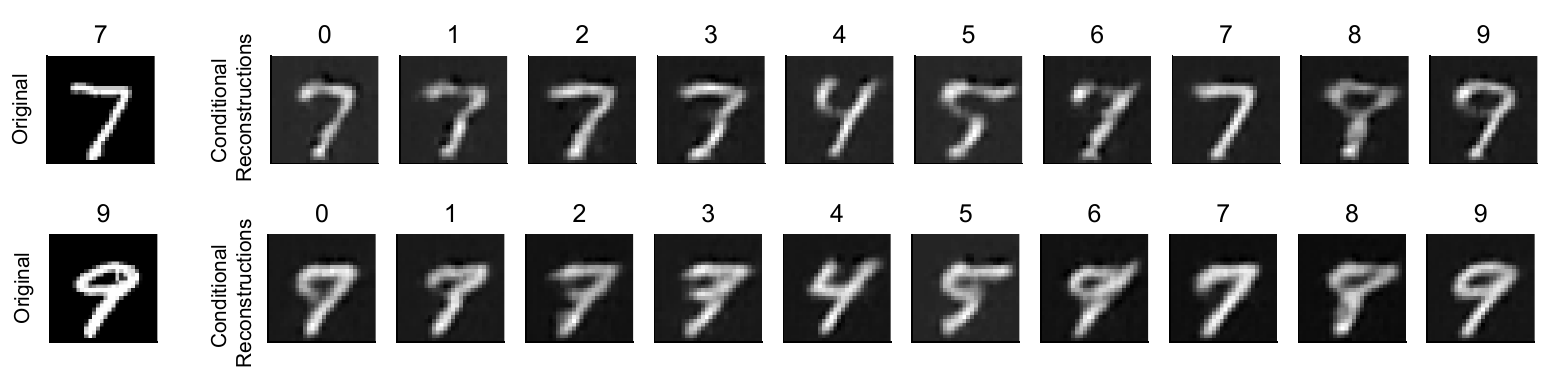}
  \caption{Conditional generation from the GBI autoencoder network. We feed an original image to the encoder and ask the decoder to reconstruct the image, as we vary the digit label input 0 to 9.}
  \label{fig:conditional_generation}
\end{figure}

The classifier used this backbone followed by a non-linear layer mapping to 10 digit classes, and then a softmax. While the GBI autoencoder produced the 8 dimensional embedding from the encoder, and concatenated the digit label one-hot vector and passed those to the decoder. Of importance, we limited the encoder embedding output to only 8 dimensional vector, otherwise the decoder would ignore the labels and use the rich, but low level visual information from the encoder. We generated a few samples from the network by providing an original image from the test set, but changing the digit label from 0 through 9, and examined the conditionally generated images, to ensure that the decoder was using the digit label (Fig \ref{fig:conditional_generation}).

\begin{figure}
    \centering
    \includegraphics[width=\textwidth]{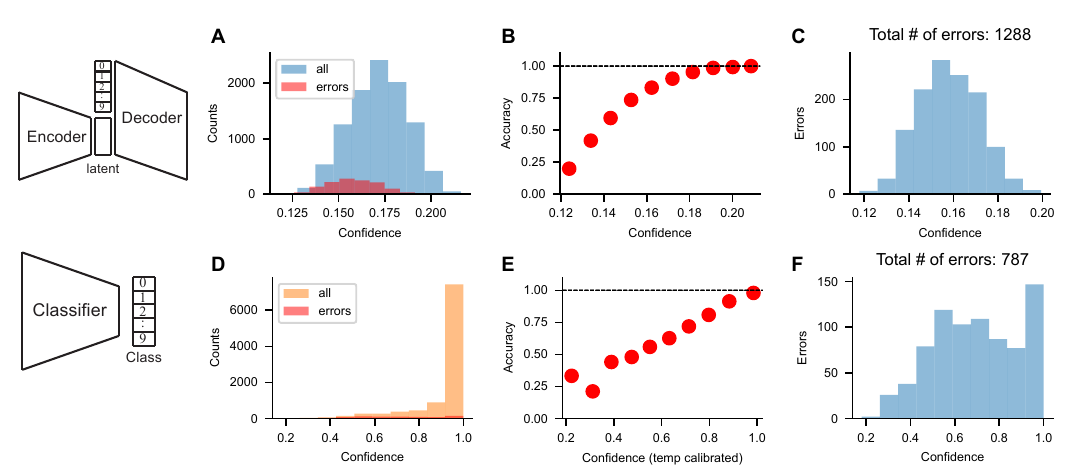}
    \caption{Confidence distributions and relationship with accuracy comparing a GBI network trained with task abstractions and a convolutional classifier.}
    \label{fig:confidence_GBI_vs_classifier}
\end{figure}

We examined the distribution of confidence values that GBI autoencoder produced, and found an informative distribution that can support post-hoc decision boundaries to get specific levels of accuracy, including 100\% accuracy in the highest confidence bins. While a classifier trained with a softmax produces a distorted distribution that does not appear informative when binned (Fig S\ref{fig:confidence_GBI_vs_classifier}). However, using the test set, one can post-hoc calibrate the softmax temperature of the classifier to obtain informative uncertainty estimates \citep{guo_calibration_2017}. 
Accordingly, our observation is that GBI does not require any post-hoc re-calibration to get an informative confidence distribution, but such as distribution is not unique.

\paragraph{Other gradient-based methods details.} First methods is norm-based inference\citep{roy_efficient_2024}. These authors use the gradients norm to infer the task. They compute the gradient norm for each possible task label and select the one with the lowest norm. Second, we compared our method to entropy-based gradient inference (EBI), following the work of \citet{wortsman_supermasks_2020}. They run a forward pass based on the superposition of all task labels each weighted by an $\alpha$ value. Subsequently, they compute the output layer activation entropy, and take a step in $\alpha$ such as to minimize that entropy. To perform a one-iteration inference, they select the supermask (or task) with the largest gradient to minimize the entropy. Given that this method is limited to cross-entropy loss we reframe image generation as minimizing a cross-entropy pixel-wise classification loss.

\begin{figure}
    \centering
    \includegraphics{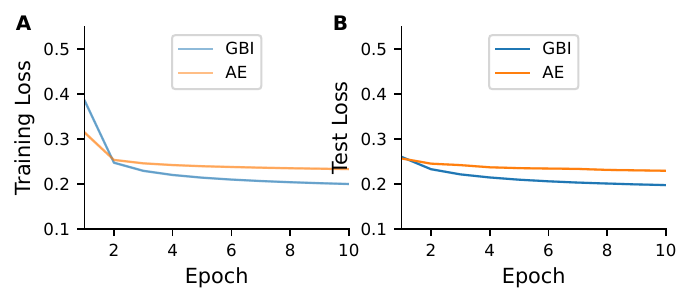}
    \caption{Comparing task-aware (GBI) and context-free autoencoder (AE) training and validation error on MNIST.}
    \label{fig:losses-gbi-vs-ae}
\end{figure}

\begin{figure}
    \centering
    \includegraphics{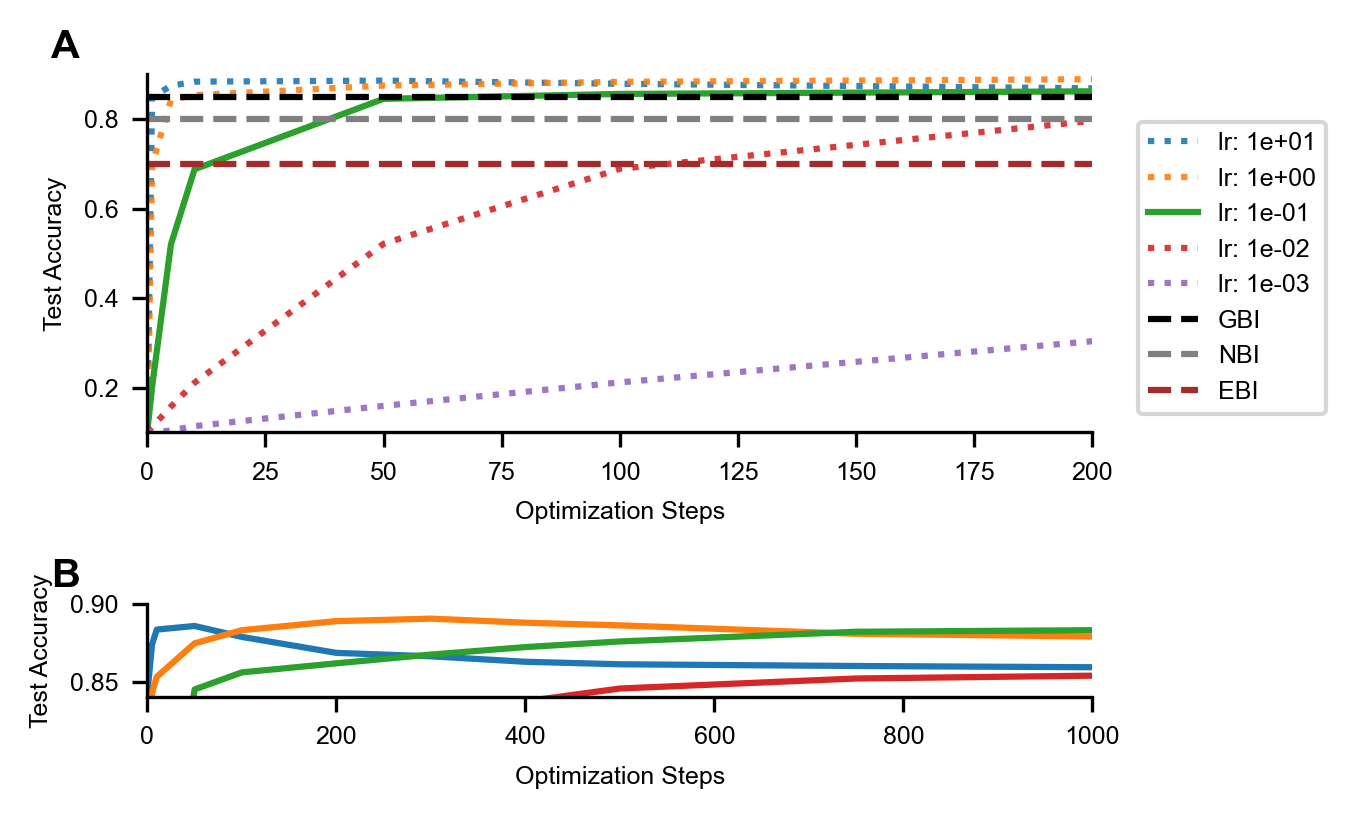}
    \caption{Variational inference iterative optimization steps vs accuracy. The models were randomly initialized in a classical variational inference fashion, and the latent representation was trained using vanilla SGD, which proved to be more stable than other optimization algorithms such as Adam and allowed control over learning rates. A) We highlight the accuracy line with the best end accuracy, other learning rates in dashed lines. Horizontal lines mark the three gradient-based methods compared in Table \ref{table:accuracies_mnist}. B) A closer look at how higher learning rates eventually lead to lower accuracy, and we picked the best learning rate based on end accuracy being the higest.}
    \label{fig:vi-performance-per-steps}
\end{figure}

\section{OOD Detection}
\label{appendix D}
\begin{figure}
    \centering
    \includegraphics[width= .9 \textwidth]{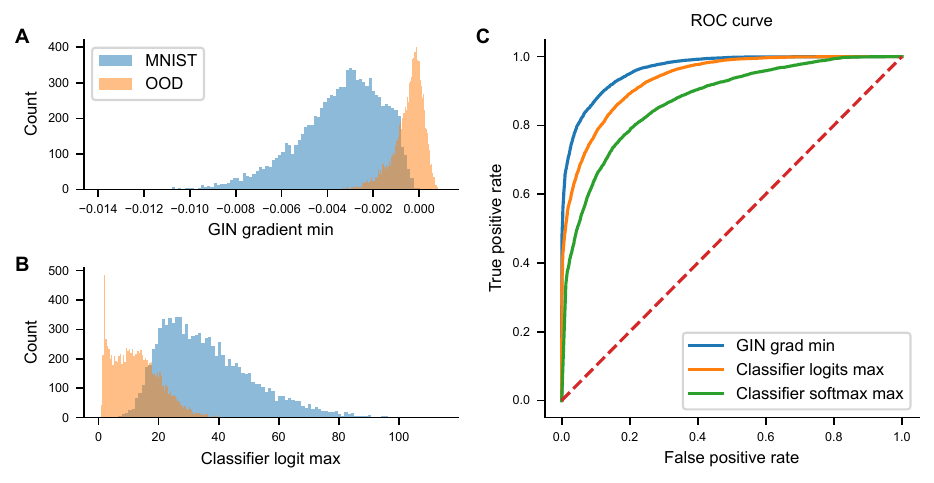}
    \caption{The figure  compares the minimum gradient values of the (A) gradient based inference network (GBI) and the maximum classifier logits when fed an in-distribution image (MNIST) versus an out-of-distribution (OOD) image (fashionMNIST). B) The OOD images have a darker average intensity compared to MNIST, resulting in lower responses in the classifier logits. Exploiting this discrepancy, a discriminating threshold based on classifier logits yields a reasonable ROC curve (C) for distinguishing between the two distributions.}
    \label{fig:ood_not_normalized}
\end{figure}

We provide the GBI gradient and classifier logit maximum values and ROC in the case where the OOD dataset was not normalized. i.e., the OOD images had a different image brightness and variance (Fig S\ref{fig:ood_not_normalized}), but we argue that an ideal image system should ignore these broad changes in illumination rather than rely on them to detect OOD samples.

\begin{table}[]
    \centering
\begin{tabular}{l c}
\hline
 & \textbf{AUCROC (Unnormalized)} \\
\textbf{Method} & \textbf{Value} \\
\hline
GBI Maxes & 0.90 $\pm$ 0.05 \\
Classifier Softmax Maxes & 0.68 $\pm$ 0.03 \\
Likelihood Regret & 1.00 $\pm$ 0.00 \\
\hline
\end{tabular}

    \caption{Same comparison as in table \ref{table:OOD_comparison}, but with the in-distribution and OOD datasets having different brightness statistics.}
    \label{table:OOD_comparison_unnormalized}
\end{table}

\section{Language Model}
\label{appendix E}




\subsection{Further experimental details on training task-aware language model}
We based our language modeling approach on the official PyTorch example for training a word-level language model, which can be found at \url{https://github.com/pytorch/examples/tree/main/word_language_model}. We largely adopted the original design choices and parameters, including the usage of a stochastic gradient descent optimizer with a decaying learning rate, a sequence length of 35 words and a batch size of 20, gradients clipped to a value of 0.25, input layer and output layer dropout rate 0.25, word-level tokenizer, and a log-softmax function at the final output layer. We used the LSTM implementation. The tokenizer's output was passed through an embedding layer, followed by concatenation with a latent representation, specifically a one-hot encoding of the dataset ID. The LSTM output passed through an MLP layer to project the LSTM's hidden units back to the number of tokens, and a log-softmax function was applied as the final output activation.

\begin{figure}
    \centering
    \includegraphics[width=.8 \textwidth]{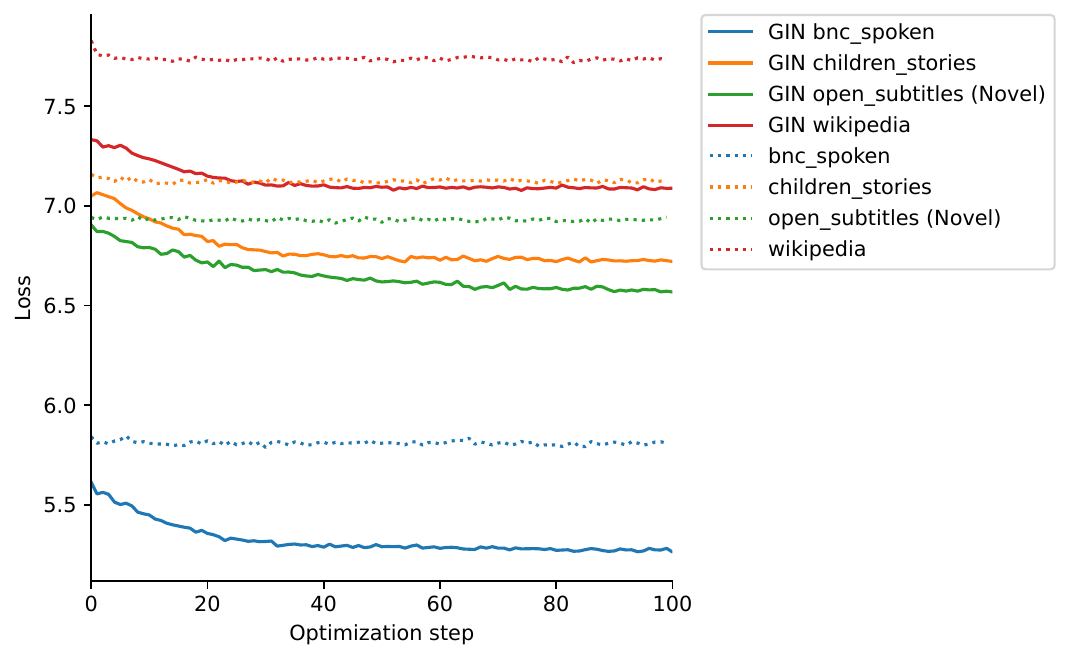}
    \caption{An example run showing the loss on predicting words from the same sequence while optimizing the latent context inputs}
    \label{fig:adaptation_example1}
\end{figure}

Note that we trained the model on 3 datasets at a time and tested generalization on a fourth dataset to keep the tokenizer from indexing more than 1 million distinct words. Adding more datasets lead to the embedding layer having a massive size that would not fit in a GPU memory.

Here we provide brief information on the datasets included in the BaybLM challenge. We use their more challenging smaller "Strict-Small" data. 

\begin{table*}[t]
    \centering
        \begin{tabular}{llr}
        \toprule
        Dataset & Domain & \textbf{Dataset Size} \\
        \midrule
        AoCHILDES \citep{macwhinney_childes_2000} & Child-directed speech & 0.44M \\
        British National Corpus (BNC),\textsuperscript{1} dialogue portion & Dialogue & 0.86M \\
        Children's Book Test \citep{hill_goldilocks_2016} & Children's books & 0.57M \\
        Children's Stories Text Corpus\textsuperscript{2} & Children's books & 0.34M \\
        Standardized Project Gutenberg Corpus \citep{lahiri_complexity_2014} & Written English & 0.99M \\
        OpenSubtitles \citep{lison_opensubtitles2016_2016} & Movie subtitles & 3.09M \\
        QCRI Educational Domain Corpus (QED; \citep{abdelali_amara_2014}) & Educational video subtitles & 1.04M \\
        Wikipedia\textsuperscript{3} & Wikipedia (English) & 0.99M \\
        Simple Wikipedia\textsuperscript{4} & Wikipedia (Simple English) & 1.52M \\
        Switchboard Dialog Act Corpus \citep{godfrey_switchboard_1992} & Dialogue & 0.12M \\
        \midrule
        \emph{Total} & -- & 9.96M \\
        \bottomrule
        \end{tabular}
    \caption{The datasets included in the BabyLM Challenge \citep{warstadt_call_2023}, please see the original paper for further details on datasets and sources. The authors reported the number of words in the training set of each corpus. \textsuperscript{1}\url{http://www.natcorp.ox.ac.uk}\ \ \ \textsuperscript{2}\url{https://www.kaggle.com/datasets/edenbd/children-stories-text-corpus}\ \ \ \textsuperscript{3}\url{https://dumps.wikimedia.org/enwiki/20221220/}\ \ \ \textsuperscript{4}\url{https://dumps.wikimedia.org/simplewiki/20221201/}}
    \label{tab:data}
\end{table*}

\begin{table}[h]
    \centering
    \begin{tabular}{lcc}
        \hline
        \textbf{Method} & \textbf{Test Accuracy (\%)} & \textbf{Model runs}\\
        \hline
        GBI & 10.92 & 1\\
        Iterative optimization  & 12.77 & 400\\
        Likelihood & 14.88 & 100\\
        \hline
    \end{tabular}
    \caption{Test results for the performance of different methods in CIFAR100 using a concatenated one-hot task abstraction to the encoder output.}
    \label{tab:cifar100-methods-results_concatenated_latent}
\end{table}





\end{document}